\documentclass[runningheads]{llncs}

\usepackage{eccv}

\usepackage{eccvabbrv}

\usepackage{graphicx}
\usepackage{booktabs}

\usepackage[accsupp]{axessibility}  % Improves PDF readability for those with disabilities.

\usepackage{multirow}
\usepackage{amsmath}
\usepackage{multicol}

\usepackage{algorithm}
\usepackage{algpseudocode}
\usepackage{xspace}

\usepackage{wrapfig}
\usepackage{graphicx}

\usepackage{hyperref}

\usepackage{orcidlink}
\renewcommand{\thefootnote}{\fnsymbol{footnote}}

\begin{document}

\title{UniTac: A Unified Multimodal Model for Cross-Sensor Tactile Understanding and Generation} 

\titlerunning{UniTac}

\author{
Jiahang Tu$^{*}$\inst{1} \and
Fengyu Yang$^{*}$\inst{2,6} \and
Chenyang Ma\inst{3} \and
Xihang Yu\inst{4} \and
Ziyao Zeng\inst{2} \and
Shaokai Wu\inst{5} \and
Hanbin Zhao$^{\dagger}$\inst{1} \and
Zhi Tao\inst{6} \and
Chao Zhang\inst{1} \and
Hui Qian\inst{1} \and
Alex Wong\inst{2}
}
\authorrunning{Tu et al.}

\institute{Zhejiang University \and Yale University \and University of Oxford \and MIT \and Shanghai Jiaotong University \and UNIX AI}

\maketitle

\begin{abstract}
    Unified multimodal models (UMMs) have shown great promise in integrating understanding and generation across diverse modalities. However, existing research rarely extends this paradigm to the tactile domain, where both object-level semantics and sensor-level configurations jointly determine the meaning of touch. To address this gap, we propose UniTac, the first UMM designed for tactile understanding and generation. UniTac models the tactile process as a transition from non-contact to contact, capturing the physical interaction between sensors and objects through a dual-level representation that encodes both sensor and object attributes. For tactile understanding, UniTac introduces two tasks, object property description and sensor identification, to enhance reasoning over physical and cross-sensor information. For tactile generation, we design a two-stage training paradigm consisting of reconstruction and alignment, together with a sensor-prior-based sampling strategy that simulates realistic tactile contact. Trained on large-scale multi-sensor datasets, UniTac achieves state-of-the-art performance in tactile understanding and generates realistic tactile signals across sensors.

{
  \renewcommand{\thefootnote}%
    {\fnsymbol{footnote}}
  \footnotetext[0]{$^*$: Equal contribution. $^\dagger$: Corresponding author.} 
}
\end{abstract}

\section{Introduction}
\label{sec:intro}

Unified multimodal models (UMMs) have recently gained significant attention for integrating perception and generation within a single architecture~\cite{li2025omniflow, shi2024lmfusion, zhou2024transfusion}.
By unifying diverse modalities, UMMs enhance the adaptability and scalability of multimodal systems, laying the foundation for developing interactive and physically grounded intelligent agents~\cite{chen2025janus, ma2025janusflow}. Within this framework, tactile understanding and generation are essential for embodied intelligence and robotic perception, as touch provides a direct means of interaction with the physical world. Accurate tactile understanding enables fine-grained manipulation~\cite{zhang2025vtla, cheng2025omnivtla}, material and surface recognition~\cite{baishya2016robust, jamali2010material}, and geometric inference beyond vision~\cite{sun2016object}. High-fidelity tactile generation facilitates simulation-based training~\cite{luu2023simulation}, strengthens cross-modal learning~\cite{tu2025texttoucher, yang2023generating}, and enables realistic tactile feedback in virtual environments~\cite{stefani2025splattouch, dou2024tactile, gao2024tactile}. Together, these capabilities strengthen multimodal perception and sensor-level grounding in embodied robotic systems.

\begin{figure}[t]
  \centering
   \includegraphics[width=1.0\linewidth]{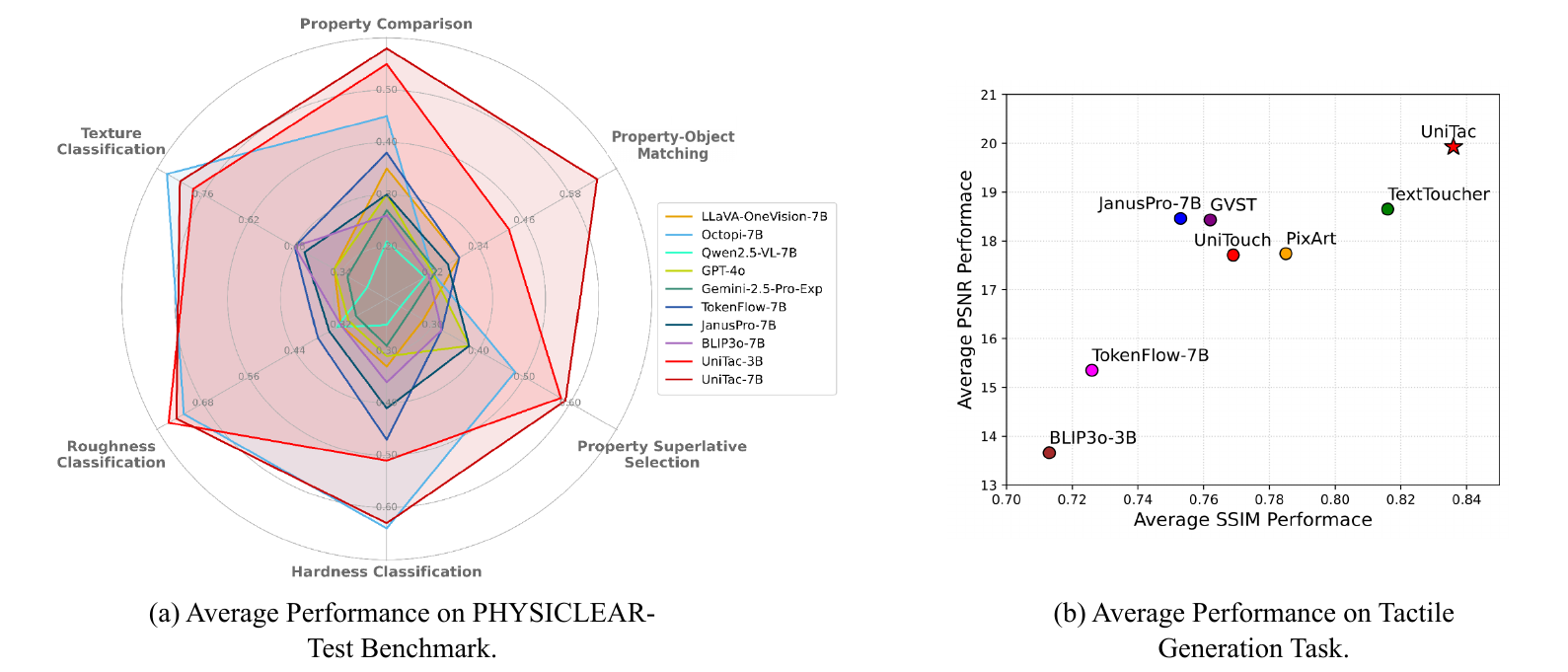}
    % \vspace{-1em} 
   \caption{Quantitative evaluation of UniTac on tactile understanding and generation tasks. (a) Average results on the PHYSICLEAR-Test benchmark across six tactile understanding tasks, where UniTac-7B achieves the strongest overall performance. (b) Average SSIM and PSNR on the tactile generation task, showing that UniTac provides superior generation quality compared with existing baselines.}
   \label{fig:lidar_benchmark}
    \vspace{-5mm} 
\end{figure}

Recent advances in visuo-tactile sensing have opened new directions for touch-centered multimodal research. Sensors such as GelSight~\cite{gel2}, DIGIT~\cite{lambeta2020digit}, and Duragel~\cite{zhang2024compact} capture tactile data in image-like formats encoding both \emph{sensor-level configurations}~\cite{tu2025texttoucher} (lighting, gel deformation, camera parameters) and \emph{object-level semantics}~\cite{gel1, gel2, lambeta2020digit} (surface geometry, hardness, roughness). However, two major challenges remain for tactile understanding. First, existing touch-language models are often trained on small, self-curated datasets, such as PHYSICLEAR~\cite{yu2024octopi} with only 482 touch videos. In contrast, the tactile research community has collectively released large-scale open datasets comprising over 400K video clips and 1.6 million frames, yet these datasets are typically used in isolation rather than being combined for joint training, which limits the representation of tactile semantics. Second, significant domain gaps among sensors hinder cross-sensor generalization due to differences in sensor characteristics and illumination conditions~\cite{tu2025texttoucher}. For tactile generation, prior work~\cite{rodriguez2024touch2touch} has explored translation between specific sensor pairs but has not addressed generalizable cross-sensor synthesis across diverse sensor types. Moreover, most tactile studies treat understanding and generation as separate problems without exploring their intrinsic connection.

Unlike visual cameras that capture a natural image in a single exposure, tactile data acquisition inherently involves two stages: a \emph{non-contact} stage capturing sensor configuration and a \emph{contact} stage recording object-level physical properties under that configuration~\cite{gel1,gel2,tu2025texttoucher}. Without object-level information, non-contact tactile data lack semantic meaning; without sensor-level information, generative models are unaware of the sensor configuration on which tactile signals should be synthesized, while understanding models cannot interpret how sensor design translates tactile patterns into object properties. For example, GelSight~\cite{gel2} encodes surface roughness through marker displacement patterns that depend on its sensor configuration. Thus, a unified multimodal model should jointly model both levels of tactile information to achieve cross-sensor tactile understanding and generation.

\begin{wrapfigure}{r}{0.6\linewidth}
\vspace{-1em}
\centering
\includegraphics[width=\linewidth]{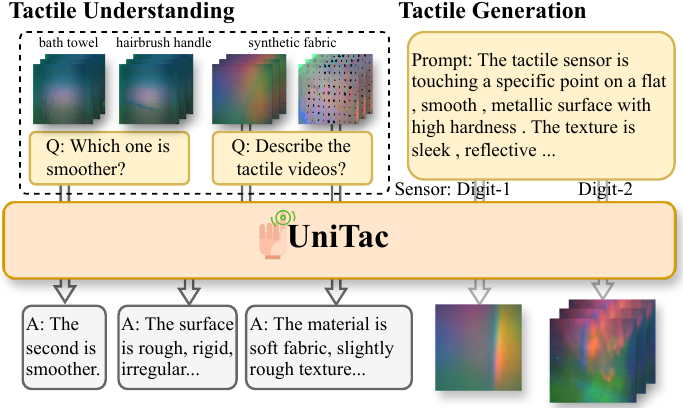}
\vspace{-1em}
\caption{Overview of UniTac for unified tactile understanding and tactile generation.}
\label{fig:unitac_overview}
\vspace{-1em}
\end{wrapfigure}

In this work, we present UniTac, the first UMM for the touch domain, designed to jointly perform tactile understanding and generation tasks within a single framework. As illustrated in Fig.~\ref{fig:unitac_overview}, UniTac unifies tactile understanding (\eg, property comparison/description) and tactile generation under a shared multimodal backbone. UniTac is trained on a large composite tactile corpus that integrates diverse open datasets comprising over 400K video clips and 1.6 million frames from multiple sensor domains. The proposed framework comprises several essential modules for tactile data processing. The touch encoder extracts meaningful representations from tactile inputs, while the multimodal large language model (MLLM) serves as a shared backbone for both understanding and generation. The sensor-aware DiT (Diffusion Transformer) projector further maps the MLLM outputs into the conditional space required by the touch decoder, which synthesizes accurate tactile data. For understanding, UniTac introduces two supervised tasks that enhance comprehension of tactile information: object property description~\cite{yu2024octopi, xie2025universal}, which captures the physical attributes of contacted objects, and sensor identification, which enables recognition of specific sensor configurations. These tasks jointly strengthen the model’s understanding of both object-level and sensor-level information. For generation, the training is divided into two stages: reconstruction and alignment. In reconstruction stage, the decoder is trained using hidden tokens from the touch encoder as conditional inputs, enabling efficient parallel training without involving the MLLM backbone. The alignment stage then trains a DiT Projector to align the MLLM’s output with encoder representations. To compensate for the absence of sensor-level cues in tactile descriptions, sensor-level tokens from the encoder are incorporated in the projector to refine the generated results. Moreover, we propose a sensor-prior-based sampling strategy that simulates the transition from non-contact to contact stages. By incorporating two level conditions into the sampling process, this method achieves flexible and accurate tactile signal generation aligned with real sensor behaviors. Extensive experiments on the PHYSICLEAR-Test benchmark~\cite{yu2024octopi} and multiple tactile datasets demonstrate strong performance of UniTac on both understanding and generation (Fig.~\ref{fig:lidar_benchmark}), compared with existing UMMs and leading tactile models. We further deploy UniTac on real robotic platforms to verify its understanding ability and the practical effectiveness of generated cross-sensor data.

Our main contributions are summarized as follows:
\begin{itemize}
    \item We introduce UniTac, the first UMM for the tactile domain, jointly modeling sensor-level configurations and object-level semantics for unified understanding and generation.
    \item We enhance tactile understanding through large-scale multi-sensor training and dual-level supervision tasks that improve object physical reasoning.
    \item We advance tactile generation via a two-stage training scheme and a sensor-prior-based sampling strategy that models the non-contact–to–contact transition for realistic tactile synthesis.
    \item We validate UniTac via extensive experiments and real-world deployment, achieving SOTA performance and demonstrating its practicality in robotic perception.
\end{itemize}

\section{Related Work}
\label{sec:related_work}

\subsection{Unified Multimodal Models}
Recent advances in unified multimodal models (UMMs) have integrated perception and generation within a single architecture~\cite{li2025omniflow, shi2024lmfusion, zhou2024transfusion, ma2025janusflow}. Works such as Janus-Pro~\cite{ma2025janusflow}, Chameleon~\cite{team2024chameleon}, and Emu3~\cite{wang2024emu3} employ Transformer-based backbones to jointly handle visual understanding and synthesis.  
Hybrid frameworks including Transfusion~\cite{zhou2024transfusion} and Show-o~\cite{xie2024show} further combine autoregressive and diffusion paradigms to exploit their complementary strengths. BLIP3-o~\cite{chen2025blip3} extends this line by introducing a diffusion transformer for CLIP-space generation with sequential training that balances understanding and generation. However, existing UMMs remain limited to visual–textual modalities. In contrast, our UniTac extends this paradigm to the tactile domain, enabling both tactile understanding and generation within a unified architecture trained on a large-scale composite dataset.

\subsection{Tactile Understanding Models}
Tactile understanding has increasingly relied on representation learning to capture contact dynamics and material properties.  
UniTouch~\cite{yang2024binding} and TVL-Link~\cite{cheng2025touch100k} align tactile inputs with visual and linguistic modalities to build consistent multimodal embeddings within a single sensor setup. AnyTouch~\cite{feng2025anytouch} proposes a cross-sensor representation unifying tactile features but focuses mainly on representation learning rather than downstream reasoning. Beyond representations, Octopi~\cite{yu2024octopi} aligns tactile videos with vision–language models for object property reasoning, while VTV-LLM~\cite{xie2025universal} explores visuo-tactile video understanding but relies on a small self-collected dataset, limiting generalization. More recent efforts such as SToLa~\cite{cheng2025stola} and CLTP~\cite{ma2025cltp} introduce commonsense reasoning and contrastive language–tactile pretraining for 3D contact understanding. Despite these advances, most models still overlook sensor-level configurations and fail to unify tactile understanding and generation within a single framework.

\subsection{Tactile Generative Models}
Tactile generation research mainly focuses on cross-modal synthesis between visual and tactile modalities. Vision2Touch~\cite{li2019connecting} establishes paired visuo-tactile data for conditional generation, and GVST~\cite{touchandgo} adopts diffusion models for high-quality bidirectional synthesis. UniTouch~\cite{yang2024binding} incorporates tactile signals into large multimodal frameworks for image-conditioned generation. Recent studies have emphasized finer control, such as TextToucher~\cite{tu2025texttoucher} for text-to-touch synthesis, ControlTac~\cite{luo2025controltac} for force- and position-controlled augmentation, and Touch2Touch~\cite{rodriguez2024touch2touch} for translation between limited sensor pairs. Nonetheless, most existing methods remain confined to single-sensor scenarios and lack unified cross-sensor modeling. UniTac overcomes these limitations by enabling multi-sensor tactile generation through the joint modeling of sensor-level configurations and object-level semantics within a unified multimodal framework.

\section{Method}
\label{sec:method}

\begin{figure}[t]
  \centering
   \includegraphics[width=0.95\linewidth]{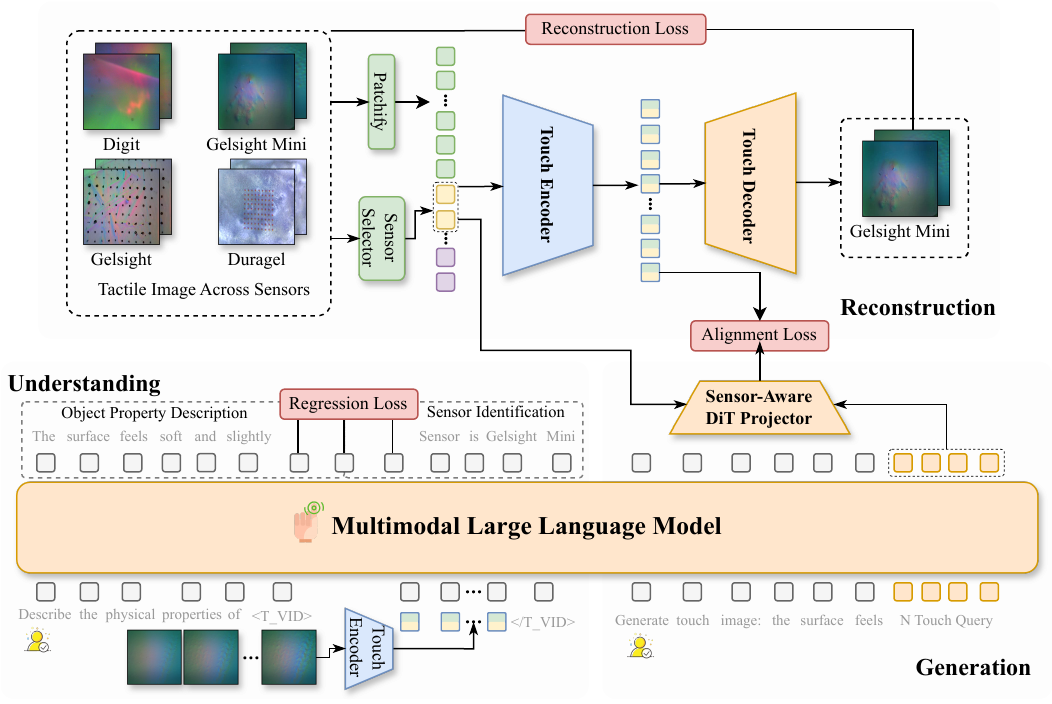}
    % \vspace{-1em} 
   \caption{Overview of the UniTac architecture. UniTac unifies tactile understanding and generation across sensors by jointly modeling sensor-level configurations and object-level semantics. The Touch Encoder extracts static and dynamic contact features, while the Multimodal Large Language Model (MLLM) integrates tactile and textual modalities for joint reasoning over object- and sensor-level information (Sec.~\ref{sec:method_dlmc}). The Sensor-Aware DiT Projector and Touch Decoder perform two-stage tactile generation (Sec.~\ref{sec:method_tsag}), combining data reconstruction with sensor-aware alignment between MLLM embeddings and tactile representations. In addition, a sensor-prior-based sampling strategy (Sec.~\ref{sec:method_spss}) models the transition from non-contact to contact, achieving realistic cross-sensor tactile synthesis.}
   \label{fig:architecture}
    \vspace{-3mm} 
\end{figure}

UniTac unifies tactile understanding and generation across sensors by jointly modeling sensor-level configurations and object-level semantics within a single framework. This formulation enables the model to reason about how different sensors perceive the same physical properties and to synthesize sensor-consistent tactile signals. To this end, we introduce three key components bridging understanding and generation. In Sec.~\ref{sec:method_dlmc}, Dual-Level Mixture Comprehension enhances tactile reasoning through object property description and sensor identification, improving cross-sensor comprehension. In Sec.~\ref{sec:method_tsag}, Two-Stage Aligned Generation advances tactile synthesis through reconstruction and sensor-aware alignment, ensuring semantic–physical coherence. In Sec.~\ref{sec:method_spss}, Sensor-Prior Sampling Strategy models the transition from non-contact to contact by embedding sensor priors during sampling, yielding realistic cross-sensor tactile generation. An overview of the full framework is shown in Fig.~\ref{fig:architecture}.

\subsection{Dual-Level Mixture Comprehension}
\label{sec:method_dlmc}
To strengthen UniTac’s reasoning in tactile understanding, we propose Dual-Level Mixture Comprehension, which jointly supervises the model from object-level and sensor-level perspectives using paired video–text data. 

Given a visuo-tactile video $V_i$ and an accompanying text $T_i$, the pretrained touch encoder produces a sequence of tactile tokens
$\mathbf{Z}_i = E_{\text{touch}}(V_i)\in\mathbb{R}^{L_v\times d}$.
We then splice these tokens into the MLLM text stream using two special markers \texttt{<T\_VID>} and \texttt{</T\_VID>}, yielding the input
$X_i = [\texttt{<T\_VID>},\ \mathbf{Z}_i,\ \texttt{</T\_VID>},\ \Pi_i,\ T_i]$, where $\Pi_i$ denotes an instructional prompt that specifies one of two comprehension objectives:
\emph{object property description} or \emph{sensor identification}.
The sequence $X_i$ is fed into the MLLM and optimized via a standard next-token prediction objective:
\vspace{-3pt}
\begin{align}
    \mathcal{L} = -\sum_{t=2+L_v+|\Pi_i|}^{|X_i|-1} \log p_\theta(x_{i,t+1} \mid x_{i,\le t}),
\end{align}
% \vspace{-3pt}
where each token in $X_i$ is trained to predict its next token $x_{i,t+1}$.  

\noindent \textbf{Object-Level Supervision (Property Description)}. Following Octopi~\cite{yu2024octopi}, we describe the contacted object using three tactile dimensions: roughness, hardness, and texture.  
For this task, $\Pi_i^{\text{prop}}$ instructs the MLLM to generate the property description (\eg, ``The surface is soft, rough and big bumps.''). By learning to predict these words conditioned on tactile embeddings, the model strengthens its grounding in object-level semantics, encouraging it to associate contact patterns with physical material properties.

\noindent \textbf{Sensor-Level Supervision (Sensor Identification)}. 
In contrast, $\Pi_i^{\text{sen}}$ guides the model to verbalize the tactile sensor identity (\eg, ``Captured by a Digit sensor.'').  
The same next-token prediction loss applies, driving the MLLM to learn sensor-related variations such as lighting, gel elasticity, and imaging resolution. This supervision enforces sensitivity to sensor-level configurations, making the representation more robust across various tactile devices.

The overall comprehension loss integrates the two tasks:
\begin{align}
    \mathcal{L}_{\text{DLMC}} = 
    \mathcal{L}_{\text{prop}} +
    \lambda_{\text{sen}}\,\mathcal{L}_{\text{sen}},
\end{align}
where $\mathcal{L}_{\text{prop}}$ and $\mathcal{L}_{\text{sen}}$ denote the respective next-token prediction losses under their prompts, and $\lambda_{\text{sen}}>0$ balance the two objectives.  
Through this dual-level supervision, UniTac learns to disentangle what changes with the \emph{object} from what changes with the \emph{sensor}, leading to more accurate tactile understanding.

\subsection{Two-Stage Aligned Generation}
\label{sec:method_tsag}

To achieve controllable and sensor-consistent tactile generation, UniTac employs a Two-Stage Aligned Generation strategy consisting of a reconstruction stage and a sensor-aware alignment stage. 

\noindent \textbf{Stage I: Reconstruction.}
The reconstruction stage focuses on learning a generative prior within the tactile domain, independent of the MLLM.  
Here, the latent representation $\mathbf{Z}_i$ extracted from the touch encoder inherently contains two types of information: $\mathbf{Z}_i = [\mathbf{Z}_i^{\text{obj}},\, \mathbf{Z}_i^{\text{sen}}]$,
where $\mathbf{Z}_i^{\text{obj}}$ encodes \emph{object-level semantics} such as hardness and roughness, and $\mathbf{Z}_i^{\text{sen}}$ represents \emph{sensor-level configurations} including illumination and gel properties. These latent tokens are fed into the touch decoder $D_{\text{touch}}$ to reconstruct the tactile signal. As this stage does not involve the MLLM backbone, it can be trained in parallel with the Dual-Level Mixture Comprehension task to improve training efficiency.

\noindent \textbf{Stage II: Sensor-Aware Alignment.}
After learning tactile priors, we align the MLLM’s textual generation space with the tactile latent space.  
Given a tactile description $T_i$ and $N$ touch queries, the MLLM produces output query embeddings: $\mathbf{\hat Q}_i = E_{\text{MLLM}}(T_i, \mathbf{Q}_i) \in \mathbb{R}^{N \times d}$, which primarily encode \textit{object-level semantics} consistent with the described physical attributes of the surface, but lack explicit sensor cues. To recover sensor awareness, we concatenate these embeddings with the specific \emph{sensor tokens} $\mathbf{S}$ obtained from the pretrained touch encoder, forming the conditional representation: $\mathbf{F}_i = [\mathbf{\hat Q}_i;\mathbf{S}]$.

% To recover sensor awareness, we concatenate these embeddings with a set of learnable \emph{sensor tokens} $\mathbf{S} = [\mathbf{s}_1,\dots,\mathbf{s}_K]$ obtained from the pretrained touch encoder, forming the conditional representation: $\mathbf{F}_i = [\mathbf{\hat Q}_i;\mathbf{S}]$.

The sensor-aware DiT projector learns a conditional vector field $v_\theta(\cdot|t,\mathbf{F}_i)$ that transforms a Gaussian prior toward the tactile latent representation $\mathbf{Z}_i$, under the guidance of $\mathbf{F}_i$. Following the rectified flow formulation~\cite{lipman2022flow, esser2024scaling}, we define a linear interpolation path between a Gaussian noise sample $\mathbf{z}\!\sim\!\mathcal{N}(0,I)$ and the target tactile latent $\mathbf{Z}_i$ as:
\begin{align}
    \mathbf{x}_t = (1-t)\,\mathbf{z} + t\,\mathbf{Z}_i, \qquad t \in [0,1]
\end{align}
whose target velocity is a constant vector
\begin{align}
 \mathbf{u}_t = \frac{d\mathbf{x}_t}{dt} = \mathbf{Z}_i - \mathbf{z}.
\end{align}
The sensor-aware DiT projector is trained by minimizing the conditional rectified flow matching loss:
\vspace{-3pt}
\begin{align}
\mathcal{L}_{\text{align}}^{\text{RF}} =
\mathbb{E}_{t\sim\mathcal{U}(0,1),\mathbf{z}\sim\mathcal{N}(0,I),\,\mathbf{Z}_i}\!
\bigl\|
v_\theta(\mathbf{x}_t|t,\mathbf{F}_i)
-
(\mathbf{Z}_i - \mathbf{z})
\bigr\|_2^2.
\end{align}
This alignment bridges the MLLM semantic output and the touch encoder representations, enabling the model to generate tactile signals that are both semantically faithful and sensor-consistent.

\subsection{Sensor-Prior Sampling Strategy}
\label{sec:method_spss}

In real tactile sensing, data acquisition follows a sequential process from a non-contact stage to a contact stage.  
The non-contact stage captures the intrinsic sensor state, while the contact stage records the physical interaction between the sensor and the object surface. Inspired by this process, we propose a Sensor-Prior Sampling Strategy (SPSS) that incorporates sensor priors into the sampling procedure, thereby simulating the gradual transition from non-contact to contact during tactile generation.

The denoised velocity field is typically estimated using classifier-free guidance (CFG)~\cite{ho2022classifier}:
\begin{align}
    \hat{v}_\theta(\mathbf{x}_t, c)
    = v_\theta(\mathbf{x}_t|t,\varnothing)
    + s\big[
    v_\theta(\mathbf{x}_t|t,c)
    - v_\theta(\mathbf{x}_t|t,\varnothing)
    \big],
\end{align}
where $c$ denotes the conditional input and $s$ is the guidance scale. However, in tactile generation, the unconditional prior $v_\theta(\mathbf{x}_t|t,\varnothing)$ does not accurately represent the non-contact initialization, as tactile signals are inherently dependent on the sensor configuration.  

To better align with the tactile data acquisition process, we replace the unconditional branch with a sensor-conditioned prior that explicitly encodes the non-contact state. The sampling process is redefined as:
\begin{align}
    \hat{v}_\theta(\mathbf{x}_t|t,\; & \mathbf{Z}_i^{\text{obj}}, \mathbf{Z}_i^{\text{sen}})
    = v_\theta(\mathbf{x}_t|t,\;\mathbf{Z}_i^{\text{sen}})
    \\ \nonumber
    & + s \big[v_\theta(\mathbf{x}_t|t,\;\mathbf{Z}_i^{\text{obj}}, \mathbf{Z}_i^{\text{sen}})
    - v_\theta(\mathbf{x}_t|t,\;\mathbf{Z}_i^{\text{sen}})
    \big].
\end{align}
Here, $v_\theta(\mathbf{x}_t|t,\mathbf{Z}_i^{\text{sen}})$ serves as the sensor prior branch, corresponding to the non-contact process, while $v_\theta(\mathbf{x}_t|t,\mathbf{Z}_i^{\text{obj}}, \mathbf{Z}_i^{\text{sen}})$ introduces the contact-specific guidance that reflects object-level semantics under the same sensor state.  
This design enforces a physically consistent transition, where the first item respects the sensor configuration and the second item incorporates contact semantics.

\section{Experiments}
\label{sec:Experiments}

\subsection{Implementation}
We train UniTac on a large-scale visuo-tactile corpus integrated from five public datasets organized by AnyTouch~\cite{feng2025anytouch}: Touch and Go~\cite{touchandgo}, Tacquad~\cite{feng2025anytouch}, TVL~\cite{fu2024touch}, SSVTP~\cite{kerr2022self}, and PHYSICLEAR~\cite{yu2024octopi}. Except for SSVTP, all datasets include paired tactile videos and textual descriptions. We preprocess and unify them into a consistent format comprising approximately 400K video clips and 1.6M frames with aligned text annotations. We filter the contact frames with tactile deformations by calculating the difference between each tactile image and the corresponding background frame in these datasets. For tactile feature extraction, the pretrained touch encoder from AnyTouch is adopted to encode both tactile images and videos. The MLLM backbone is based on Qwen-VL 2.5~\cite{wang2024qwen2}, and the touch decoder is implemented with SANA~\cite{xie2024sana}. Additional implementation details are provided in Appendix~\ref{sec:Implementation Details}.

For tactile understanding, we evaluate simple perceptual recognition through hardness, roughness, and texture classification, and complex reasoning through Property Comparison (PC), Property-Object Matching (POM), and Property Superlative Selection (PSS) on the PHYSICLEAR-Test benchmark~\cite{yu2024octopi}. For tactile generation, following GVST~\cite{touchandgo} and TextToucher~\cite{tu2025texttoucher}, we evaluate synthesis quality using SSIM and PSNR by randomly sampling 60K generated tactile images and comparing them with their corresponding ground-truth samples.

\begin{figure}[t]
  \centering
   \includegraphics[width=1.0\linewidth]{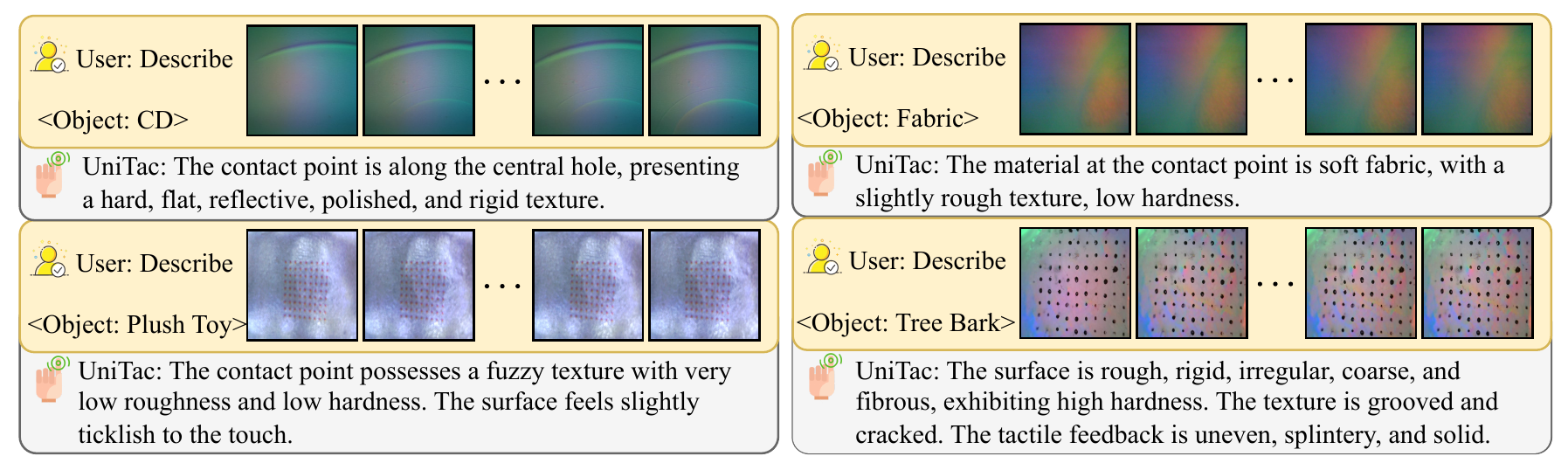}
    % \vspace{-1.5em} 
   \caption{Object property description of tactile videos across various tactile sensors. UniTac generates object-aware tactile descriptions that align with physical properties of the contacted materials.}
   \label{fig:touch_video_understanding_opd_res}
    \vspace{-2mm} 
\end{figure}

\begin{figure}[t]
  \centering
   \includegraphics[width=1.0\linewidth]{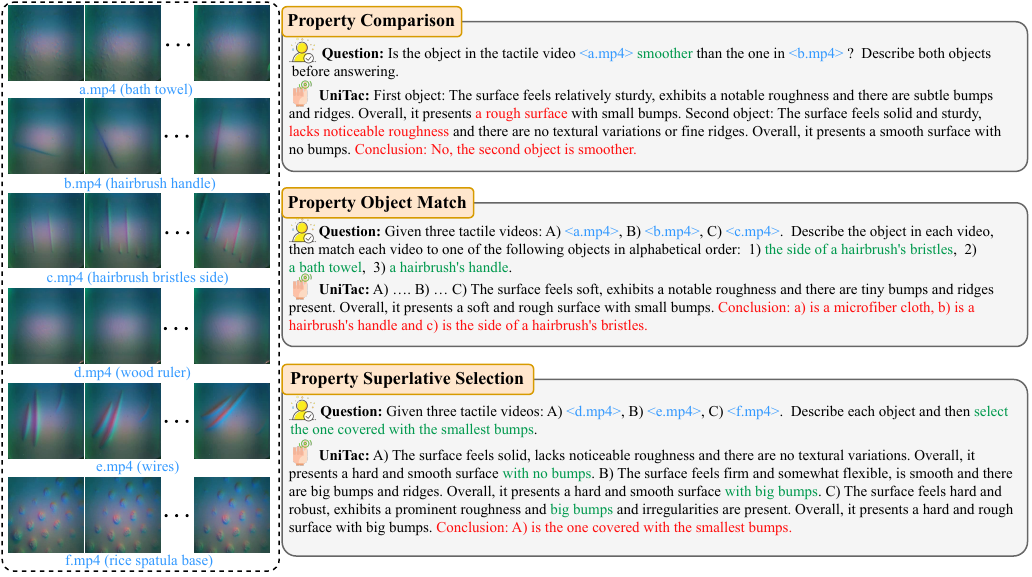}
    \vspace{-1em} 
   \caption{Tactile video understanding across various understanding tasks. We evaluate UniTac on three representative understanding tasks: Property Comparison, Property–Object Matching, and Property Superlative Selection. UniTac first provides fine-grained descriptions of surface attributes (\eg, roughness, bumps, hardness) and then performs reasoning to derive the final answer. The results show that UniTac can distinguish tactile differences through multi-step tactile reasoning capability.}
   \label{fig:touch_video_understanding_res}
    \vspace{-1em} 
\end{figure}

\subsection{Tactile Understanding}

\begin{table}[t]
    \centering
    \caption{Comparing the understanding capability of UniTac with other generative and unified multimodal models. UniTac-7B achieves the best average performance across all metrics, indicating stronger fine-grained tactile reasoning and understanding on PHYSICLEAR-Test benchmark. PC refers to Property Comparison task; POM refers to Property–Object Matching task; PSS refers to Property Superlative Selection task.}
    % \vspace{-0.5em}
    \label{tab:und_benchmark}
    \resizebox{1.0\textwidth}{!}{
        \begin{tabular}{l|l|c|cccccc|c}
        \toprule
    	Type & Model & Method & \textbf{PC}$\uparrow$ & \textbf{POM}$\uparrow$ & \textbf{PSS}$\uparrow$ & \textbf{Hardness}$\uparrow$ & \textbf{Roughness}$\uparrow$ & \textbf{Texture}$\uparrow$  & \textbf{Average}$\uparrow$ \\
    	\midrule
            \multirow{5}{*}{Und. Only} 
            & GPT-4o~\cite{hurst2024gpt} & AR & 30.87 & 22.62 & 38.05 & 31.37 & 30.48 & 36.51 & 31.65 \\
            & Qwen2.5-VL-7B~\cite{bai2025qwen2} & AR & 21.47 & 19.92 & 23.05 & 25.73 & 33.33 & 26.62 & 25.01 \\
            & Gemini-2.5-Pro-Exp~\cite{comanici2025gemini} & AR & 26.71 & 23.47	& 25.06	& 29.19	& 28.44	& 32.18	& 27.50 \\
            & LLaVA-OneVision-7B~\cite{li2024llava} & AR & 35.42 & 29.11 & 28.44 & 33.33 & 32.18 & 36.51 & 32.49 \\
            & Octopi-7B~\cite{yu2024octopi} & AR & 45.50 & 22.22 & 48.00 & \textbf{64.10} & 73.92	& 87.17	& 57.31 \\
            \midrule
            \multirow{5}{*}{Und. and Gen.}
            & TokenFlow-7B~\cite{qu2025tokenflow} & AR & 38.05 & 29.48 & 32.56 & 47.68 & 38.70 & 48.31 & 39.13 \\ 
            & JanusPro-7B~\cite{chen2025janus} & AR + Diff & 30.18 & 26.71 & 38.05 & 41.63 & 35.82 & 45.37 & 36.29 \\
            & BLIP3o-3B~\cite{chen2025blip3} & AR+ Diff & 26.71 & 21.47 & 32.18 & 36.51 & 32.18 & 48.00 & 32.84 \\
            & UniTac-3B(Ours) & AR + Diff & 54.97 & 42.13 & 58.90 & 51.28 & \textbf{76.92} & 79.48 & 60.61 \\
            & UniTac-7B(Ours) & AR + Diff & \textbf{57.30} & \textbf{64.61} & \textbf{59.22} & 61.53 & 74.35 & \textbf{82.05} & \textbf{66.51} \\
        \bottomrule
        \end{tabular}}
    % \vspace{-1em}
\end{table}

To evaluate the tactile understanding capability of UniTac, we compare it with advanced multimodal models like Qwen2.5-VL~\cite{bai2025qwen2} and GPT-4o~\cite{hurst2024gpt}. We further include representative UMMs that unify understanding and generation. The quantitative results are summarized in Tab.~\ref{tab:und_benchmark}. UniTac-7B achieves the highest overall score of 66.51, surpassing all previous models.  
Notably, the overall improvement mainly comes from UniTac's stronger tactile reasoning capability rather than uniform gains across all tasks. On the three reasoning-oriented tasks, i.e., PC, POM, and PSS, UniTac-7B improves over the strongest UMM baseline Octopi-7B by 11.80, 42.39, and 11.22 points, respectively. These results demonstrate that UniTac can better ground tactile observations into high-level physical and semantic concepts, leading to more reliable reasoning over material properties and tactile interactions.

In Fig.~\ref{fig:touch_video_understanding_opd_res}, we present qualitative results of the object property description (OPD) task, which is explicitly used during training to enhance tactile understanding. Across diverse tactile sensors, UniTac generates coherent and physically grounded descriptions that align with material properties and sensor configurations. For instance, it correctly characterizes the rigid, polished surface of a CD and distinguishes it from the soft, fuzzy texture of a plush toy, reflecting accurate reasoning over hardness and roughness cues. Building upon this capability, Fig.~\ref{fig:touch_video_understanding_res} further evaluates UniTac on complex tactile reasoning tasks. Rather than directly predicting answers, the model implicitly relies on fine-grained surface attributes, such as roughness, bump distribution, and stiffness, to support comparison, matching, and superlative selection.

\subsection{Tactile Generation}

\begin{table}[t]
    \caption{Comparing the generation capability of UniTac with other generative and unified multimodal models. Results are reported across four tactile sensors (Digit, GelSight, GelSight Mini, and Duragel). UniTac achieves the highest average SSIM and PSNR among unified multimodal models and remains competitive with generation-only methods, demonstrating strong cross-sensor tactile synthesis capability.}
    % \vspace{-0.5em}
    \label{tab:gen_benchmark}
    \resizebox{\textwidth}{!}{
    \begin{tabular}{l|l|cccccccc|cc}
    \toprule
    \multirow{2}{*}{Type}          & \multirow{2}{*}{Model} & \multicolumn{2}{c}{\textbf{Digit}} & \multicolumn{2}{c}{\textbf{Gelsight}} & \multicolumn{2}{c}{\textbf{Gelsight Mini}} & \multicolumn{2}{c}{\textbf{Duragel}} & \multicolumn{2}{|c}{\textbf{Average}} \\
    % \cmidrule(lr){3-10}
                                   & & \textbf{SSIM}$\uparrow$ & \textbf{PSNR}$\uparrow$ & \textbf{SSIM}$\uparrow$ & \textbf{PSNR}$\uparrow$ & \textbf{SSIM}$\uparrow$ & \textbf{PSNR}$\uparrow$ & \textbf{SSIM}$\uparrow$ & \textbf{PSNR}$\uparrow$ & \textbf{SSIM}$\uparrow$ & \textbf{PSNR}$\uparrow$ \\
    \midrule
    \multirow{4}{*}{Gen. Only}     & GVST~\cite{yang2023generating} & 0.841 & 18.72 & 0.655 & 14.58 & 0.883 & 18.90 & 0.429 & 14.98 & 0.762 & 18.43 \\
                                   & UniTouch~\cite{yang2024binding} & 0.859 & 18.17 & 0.636 & 14.18 & 0.851 & 18.63 & 0.440 & 14.31 & 0.769 & 17.71 \\
                                   & PixArt-$\alpha$~\cite{chen2023pixart} & 0.877 & 18.10 & 0.641 & 14.09 & 0.869 & 18.77 & 0.446 & 14.36 & 0.785 & 17.74 \\
                                   & TextToucher~\cite{tu2025texttoucher} & 0.901 & 20.90 & 0.662 & 15.42 & 0.896 & 20.38 & \textbf{0.473} & 16.36 & 0.816 & 18.65 \\
    \midrule
    \multirow{5}{*}{Und. and Gen.} & TokenFlow-7B~\cite{qu2025tokenflow} & 0.834 & 17.47 & 0.628 & 14.04 & 0.758 & 15.67 & 0.431 & 12.72 & 0.726 & 15.35 \\
                                   & JanusPro-7B~\cite{chen2025janus} & 0.854 & 20.18 & 0.653 & 15.34 & 0.894 & 19.67 & 0.464 & 17.73 & 0.753 & 18.46 \\
                                   & BLIP3o-7B~\cite{chen2025blip3} & 0.821 & 16.30 & 0.618 & 13.46 & 0.746 & 14.74 & 0.405 & 10.92 & 0.713 & 13.66 \\
                                   & UniTac & \textbf{0.915} & \textbf{21.26} & \textbf{0.683} & \textbf{16.28} & \textbf{0.946} & \textbf{24.56} & 0.472 & \textbf{18.44} & \textbf{0.836} & \textbf{19.93} \\
                                   % & UniTac-7B &  &  &  &  &  &  &  &  &  & \\
    \bottomrule
    \end{tabular}}
% \vspace{-1em}
\end{table}

\begin{figure}[t]
  \centering
   \includegraphics[width=1.0\linewidth]{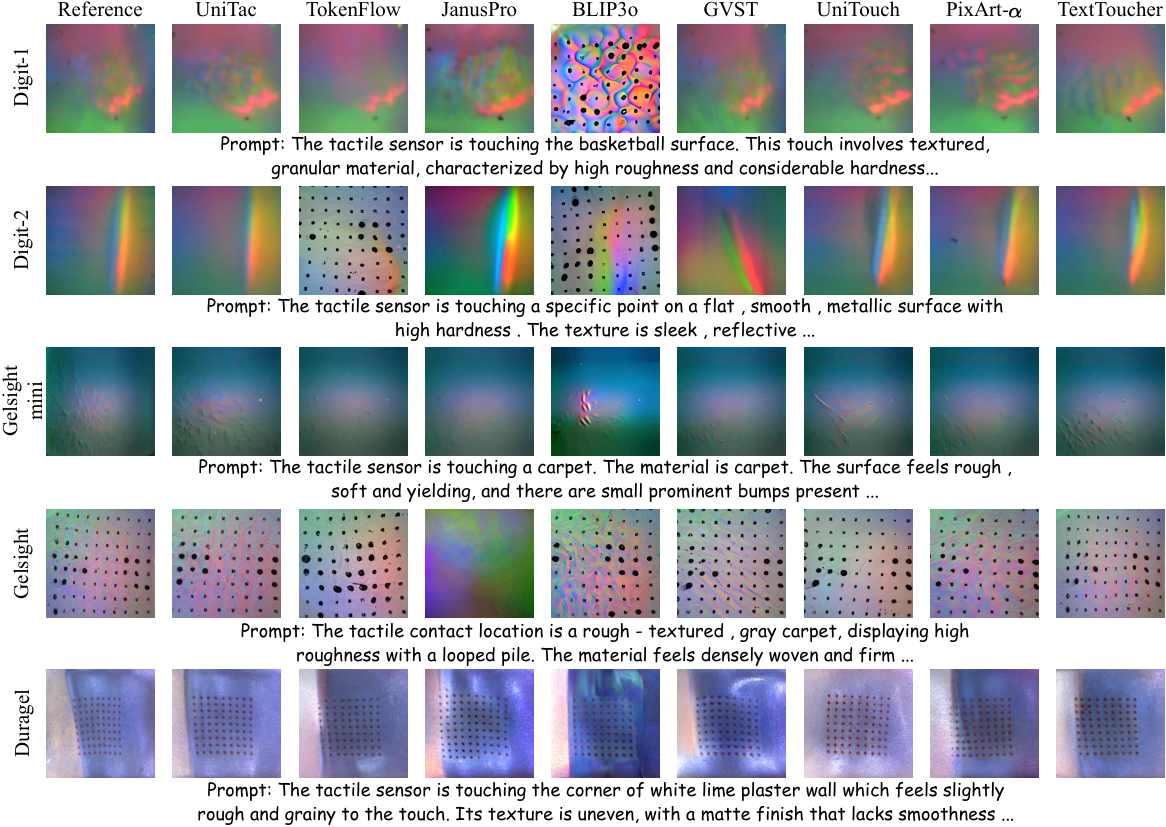}
    % \vspace{-2em} 
   \caption{Qualitative comparison of tactile image generation across various tactile sensors. UniTac consistently generates realistic and physically coherent tactile images across diverse sensors and configurations.}
   \label{fig:touch_image_generation_res}
    \vspace{-1.5em} 
\end{figure}

\noindent \textbf{Image Generation.}
We evaluate UniTac’s tactile image generation on four representative sensors: Digit, GelSight, GelSight Mini, and Duragel. As shown in Tab.~\ref{tab:gen_benchmark}, UniTac achieves the highest average performance of 0.836 SSIM and 19.93 PSNR, surpassing all state-of-the-art generative and unified models. We also observe that all models perform relatively lower on Duragel due to unstable acquisition conditions and significant variation in sensor calibration and gel state. The collected tactile images are often blurrier and less consistent in lighting and contact geometry, increasing intra-domain variance and making it harder for models to learn stable patterns between tactile appearance and physical properties, leading to overall lower metrics across all methods.

In Fig.~\ref{fig:touch_image_generation_res}, UniTac consistently generates realistic and physically coherent tactile images across different sensor types and configurations. It can also accurately reflect distinct sensor configuration even the same sensor type, such as the results in Digit-1 and Digit-2. Unlike other UMMs that misrepresent sensor states or lose spatial fidelity, UniTac preserves fine structural and illumination cues, demonstrating robust cross-sensor consistency enabled by its joint modeling of sensor- and object-level information.

\begin{figure}[t]
  \centering
   \includegraphics[width=1.0\linewidth]{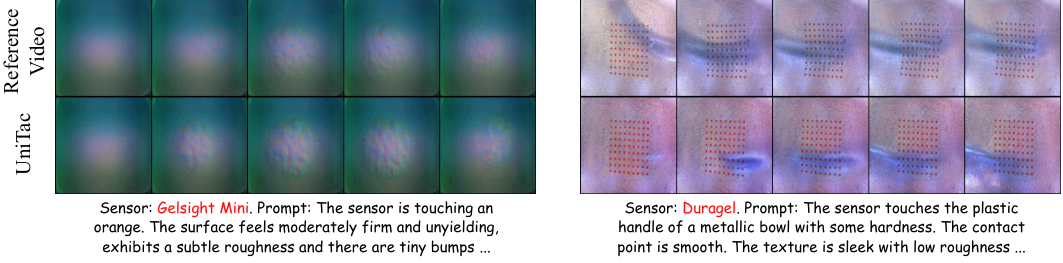}
    % \vspace{-1em} 
   \caption{Cross-sensor tactile video generation results. Left (GelSight Mini, orange): UniTac reproduces the fine, bumpy texture and progressive deformation consistent with the orange peel surface. Right (Duragel, bowl handle): UniTac generates smooth contact patterns with localized indentation and realistic dynamic changes throughout the contact process.}
   \label{fig:touch_video_generation_res}
    % \vspace{-1em} 
\end{figure}

\noindent \textbf{Video Generation.}
Beyond static image synthesis, we extend UniTac to tactile video generation by replacing the touch decoder with Wan v2.2~\cite{wan2025wan}, while keeping the same sensor-aware conditioning mechanism. This design enables temporally coherent synthesis without altering the unified multimodal backbone or the sensor-specific conditioning strategy. As shown in Fig.~\ref{fig:touch_video_generation_res}, UniTac generates continuous tactile sequences across different sensors, capturing both the onset of contact and the gradual pressure propagation over time. The generated videos preserve fine-grained texture cues (\eg, the granular surface of an orange) as well as sensor-specific deformation patterns and illumination dynamics. Besides, UniTac maintains temporal consistency in dot-grid displacement and contact region evolution. These results demonstrate that the proposed framework generalizes naturally from static tactile image synthesis to dynamic tactile video modeling, enabling realistic simulation of contact processes under diverse sensor configurations. More results are provided in Appendix~\ref{sec:qualitative}.

\subsection{Ablation Studies}

To examine the contribution of each component in UniTac, we conduct a series of ablation experiments, as summarized in Tab.~\ref{tab:ablation}. More ablation studies are provided in Appendix~\ref{sec:more_results}. Removing the Sensor Identification objective causes a clear drop on PHYSICLEAR (from 60.61 to 57.38), indicating that explicitly modeling sensor-level configurations substantially enhances the UMM’s ability to interpret tactile data across various tactile sensors. Without the Dual-Level Comprehension, the performance decreases sharply across all metrics, as the MLLM degenerates into a standard Qwen-VL model lacking tactile understanding. This confirms that tactile comprehension not only improves reasoning performance but also provides critical semantic priors for generation. In contrast, removing the DiT Projector or Sensor-Prior Sampling, which are trained independently from the MLLM, has limited impact on understanding but reduces generative fidelity. This reflects that sensor-level representation remains essential for maintaining structural realism and physical consistency in generation.

\begin{table}[t]
\caption{Performance impact of different modules in UniTac through ablation study.}
% \vspace{-1em}
\centering
\label{tab:ablation}
\setlength{\tabcolsep}{18pt} 
\resizebox{0.95\linewidth}{!}{
\begin{tabular}{lccc}
\toprule
Component & PHYSICLEAR & SSIM  & PSNR  \\ \midrule
UniTac & \textbf{60.61} & \textbf{0.836} & \textbf{19.93} \\
\emph{w/o} Sensor Identification & 57.38 & 0.822 & 19.91 \\
\emph{w/o} Dual-Level Comprehension & 26.52 & 0.758 & 18.14 \\
\emph{w/o} DiT Projector & 60.61 & 0.794 & 19.25 \\
\emph{w/o} Sensor-Prior Sampling & 60.61 & 0.817 & 19.49 \\ \bottomrule
\end{tabular}
}
\vspace{-1em}
\end{table}

\subsection{Real-World Deployment and Validation}
\label{sec:real_world_deployment}

To evaluate the practicality of UniTac, we deploy UniTac in a real robotic platform and validate both its understanding capability and the utility of generated tactile data.  

\begin{figure}[t]
  \centering
   \includegraphics[width=1.0\linewidth]{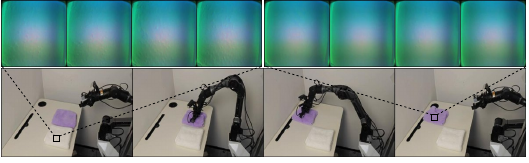}
    \vspace{-2em} 
   \caption{The robot compares two visually similar fabrics through the Property Comparison task, identifies the smoother one as more suitable for baby skin contact. Tactile differences between the two materials are magnified for clarity.}
\label{fig:touch_video_understanding_real_environ}
    % \vspace{-4mm} 
\end{figure}

\noindent \textbf{Understanding in Real Interaction.}
We apply UniTac in a real-world grasping scenario involving material selection for delicate skin contact, aiming to identify the fabric most suitable for wiping an infant's skin. This task requires subtle tactile discrimination beyond visual similarity. As shown in Fig.~\ref{fig:touch_video_understanding_real_environ}, given two visually similar fabrics with distinct tactile properties, UniTac analyzes the tactile feedback and performs fine-grained comparison of surface attributes. The model identifies the purple towel as exhibiting lower roughness and a smoother texture. Based on this tactile comparison outcome, the robotic system selects and grasps the smoother fabric as the preferred option for baby skin contact.

We further quantify this setting as a language-guided tactile fabric selection and grasping task. A rollout is considered fully successful only when the robot selects the baby-suitable fabric, grasps a single-layer edge or corner, lifts it stably, and avoids visible slipping. As shown in Tab.~\ref{tab:towel_selection_results}, the RGB-only VLA baseline can often localize the fabric regions visually, but it struggles with tactile-semantic selection and single-layer contact-state estimation. In contrast, VTLA-real achieves the best performance by directly observing tactile cues such as softness, contact area, and slip tendency. VTLA-predict, while not using real-time tactile input during inference, still substantially outperforms VLA, suggesting that predicted tactile representations preserve useful physical cues for contact-aware manipulation. Additional deformation-aware grasping results are provided in Appendix~\ref{sec:appendix_robot_tactile_utility}.

\begin{table}[t]
\centering
\caption{Rollout results for language-guided tactile towel selection and grasping. VLA is an RGB-only baseline that maps visual observations, robot state, and language instructions directly to actions. VTLA-real additionally uses real GelSight tactile observations during rollout. VTLA-pred predicts tactile representations from RGB/state inputs at inference, testing whether tactile-aware control can be achieved without real-time tactile sensing. Implementation details are provided in Appendix~\ref{sec:appendix_robot_tactile_utility}.}
\label{tab:towel_selection_results}
\setlength{\tabcolsep}{5mm}
\resizebox{0.89\linewidth}{!}{
\begin{tabular}{l|ccc}
\toprule
Method & Selection Success $\uparrow$ & Grasping Success $\uparrow$ & Overall Success $\uparrow$ \\
\midrule
VLA & 11/20 (55\%) & 6/20 (30\%) & 4/20 (20\%) \\
VTLA-real & 20/20 (100\%) & 19/20 (95\%) & 19/20 (95\%) \\
VTLA-pred & 18/20 (90\%) & 16/20 (80\%) & 16/20 (80\%) \\
\bottomrule
\end{tabular}}
\end{table}

\begin{table}[t]
\centering
\caption{Effectiveness of UniTac-generated GelSight data in improving cross-sensor grasp classification performance. By augmenting with UniTac-generated GelSight data, cross-sensor grasp accuracy improves while maintaining high Digit performance.}
\vspace{-0.5em}
\label{tab:generated_cls}
\setlength{\tabcolsep}{3mm}
\resizebox{0.7\linewidth}{!}{
\begin{tabular}{lcc}
\toprule
Data                  & Digit Grasp(\%) & Gelsight Grasp(\%) \\ \midrule
Digit                 & 98.89       & 50.00          \\
Digit+UniTac-Gelsight & 99.07       & 99.37          \\ \bottomrule
\end{tabular}
}
\vspace{-1em}
\end{table}

\noindent \textbf{Effectiveness of Generated Tactile Data.}
Tactile sensing hardware evolves rapidly, making large-scale data recollection for each newly introduced device both costly and time-consuming. This challenge is particularly pronounced in cross-sensor scenarios, where models trained on one sensor often fail to generalize to another due to differences in illumination patterns, marker layouts, and deformation characteristics. UniTac alleviates this issue by generating sensor-specific tactile data conditioned on object and contact information, enabling efficient cross-domain adaptation without additional real data collection. We evaluate this capability in a cross-sensor grasp classification task. In Tab.~\ref{tab:generated_cls}, when trained only on Digit data, the classifier achieves strong in-domain performance (98.89\%) but suffers a severe drop when tested on GelSight (50.00\%), highlighting the substantial sensor gap. By augmenting the training set with UniTac-generated GelSight samples, the GelSight accuracy improves to 99.37\%, while Digit performance remains stable (99.07\%). This suggests that UniTac can serve as a scalable data generation framework to support rapid adaptation for emerging tactile devices.

\section{Conclusion}
\label{sec:conclusion}

This paper introduces UniTac, the first unified multimodal model (UMM) designed for cross-sensor tactile understanding and generation. UniTac bridges the gap between tactile reasoning and synthesis by jointly modeling sensor-level configurations and object-level semantics through a dual-level mixture comprehension framework. A two-stage aligned generation paradigm with a sensor-prior sampling strategy further ensures physically consistent tactile synthesis. Extensive experiments demonstrate that UniTac outperforms existing UMMs and specialized tactile models in both understanding and generation, highlighting its potential to advance unified multimodal modeling.

\paragraph{\textbf{Acknowledgments}.}
This work was supported in part by the National Natural Science Foundation of China under Grant No. 62402430. Hanbin Zhao was also partially supported by the Zhejiang Provincial Natural Science Foundation of China under Grant No. LQN25F020008. Alex Wong was supported by the NSF Athena AI Institute under Grant No. 2112562 and by the Global Industrial Technology Cooperation Center (GITCC) through a grant agreement with the Korea Institute for Advancement of Technology (KIAT), Project No. P0028922.

\bibliographystyle{splncs04}
\bibliography{main}

\clearpage
\setcounter{page}{1}
% \maketitlesupplementary
\appendix

\setcounter{table}{0}  
\setcounter{figure}{0}
\setcounter{equation}{0}

\renewcommand{\thetable}{\Alph{table}}
\renewcommand{\thefigure}{\Alph{figure}}
\renewcommand{\theequation}{\Alph{equation}}

\noindent\textbf{Overview.} In this supplementary material, we submit the source code in the ``UniTac'' folder and provide more experiments about our method. In Section~\ref{sec:Implementation Details}, we provide the specifics of our experiment setup. We then offer a deeper analysis of the parameter settings and present insightful findings in Section~\ref{sec:more_results}. More samples of our method are provided in Section~\ref{sec:qualitative} and preliminaries of  rectified flow matching are presented in Section~\ref{sec:preliminaries}.

\section{Implementations Details}
\label{sec:Implementation Details}

\newcommand{\TVID}{\texttt{<T\_VID>}\xspace}
\newcommand{\ETouch}{E_{\text{touch}}}
\newcommand{\DTouch}{D_{\text{touch}}}
\newcommand{\EMLLM}{E_{\text{MLLM}}}

\subsection{Model Architecture}
UniTac is composed of four key modules: a Touch Encoder, a Multimodal Large Language Model (MLLM) backbone, a Sensor-Aware DiT Projector, and a Touch Decoder for tactile image and video generation.

\paragraph{Touch Encoder.}
The touch encoder adopts the AnyTouch~\cite{feng2025anytouch} architecture with a ViT-B/16 backbone pretrained on large-scale multi-sensor tactile videos and images. This pretraining allows the encoder to effectively extract both spatial and temporal features that generalize across heterogeneous tactile sensors (e.g., Digit, GelSight, and Duragel). In addition, the encoder incorporates a set of learnable sensor tokens, where each sensor type is associated with five tokens to encode sensor-specific characteristics. The encoder outputs 768-dimensional latent tokens containing object-level semantics and sensor-level configurations.

\paragraph{MLLM Backbone.}
We employ two configurations of the MLLM based on Qwen-VL 2.5: a lightweight 3B model and a larger 7B model for tactile understanding. The 3B backbone already achieves strong performance on tactile understanding tasks, offering efficient inference and better generalization. Therefore, we adopt the 3B variant as the backbone for tactile generation to balance quality and computational efficiency. And our tactile image generation model comprises approximately 5B parameters. In both configurations, the visual projector of Qwen-VL is replaced with a tactile embedding adaptor that seamlessly integrates tactile latent tokens into the textual sequence for joint reasoning.

\paragraph{Sensor-Aware DiT Projector.}
For mapping the MLLM textual outputs into the tactile latent space, we employ NextDiT~\cite{zhuo2024lumina}, a 24-layer diffusion transformer designed for efficient conditional feature alignment~\cite{chen2025blip3}. The projector integrates semantic embeddings from the MLLM with sensor priors extracted by the touch encoder, learning a conditional velocity field that continuously aligns textual and tactile representations. This alignment enables cross-sensor tactile generation that remains physically consistent with real sensor behaviors.

\paragraph{Touch Decoder.}
For tactile image generation, we adopt the SANA~\cite{xie2024sana} diffusion architecture, capable of synthesizing 512 × 512 high-fidelity tactile images conditioned on sensor-aware latent embeddings.
For tactile video generation, we replace the image decoder with Wan v2.2~\cite{wan2025wan}, which produces 13-frame sequences at 448 × 448 resolution while preserving temporal continuity of contact dynamics.

\subsection{Training Setup and Environment}
\label{sec:training}
UniTac training consists of three major stages: reconstruction, dual-level comprehension, and sensor-aware alignment. The reconstruction and dual-level comprehension stages are trained in parallel, as the former does not involve the MLLM backbone required by the latter.

\begin{algorithm}[h]
\caption{Dual-Level Mixture Comprehension (DLMC) Training}
\label{alg:dlmc}
\begin{algorithmic}[1]
\Require Tactile video $V_i$, paired text $T_i$, property prompt $\Pi_i^{\text{prop}}$, sensor prompt $\Pi_i^{\text{sen}}$, weighting factor $\lambda_{\text{sen}}$
\State $Z_i \gets \ETouch(V_i)$ \Comment{encode tactile video into token sequence}
\State $X_i^{\text{prop}} \gets [\TVID, Z_i, \texttt{</T\_VID>}, \Pi_i^{\text{prop}}, T_i]$
\State $X_i^{\text{sen}}  \gets [\TVID, Z_i, \texttt{</T\_VID>}, \Pi_i^{\text{sen}},  T_i]$
\Function{NextTokenLoss}{$X$, $T$}
  \State Identify the starting index $t_{\text{txt}}$ of text tokens $T$ in $X$
  \State $L \gets 0$
  \For{$t = t_{\text{txt}}$ \textbf{to} $|X|-1$}
    \State $L \gets L - \log p_{\theta}(x_{t+1} \mid x_{\le t})$ \Comment{predict only textual tokens, Eq.~(1)}
  \EndFor
  \State \Return $L$
\EndFunction
\State $L_{\text{prop}} \gets \Call{NextTokenLoss}{X_i^{\text{prop}}, T_i}$
\State $L_{\text{sen}} \gets \Call{NextTokenLoss}{X_i^{\text{sen}}, T_i}$
\State $L_{\text{DLMC}} \gets L_{\text{prop}} + \lambda_{\text{sen}} L_{\text{sen}}$ \Comment{Eq.~(2)}
\State Update parameters $\theta$ by gradient descent on $L_{\text{DLMC}}$
\end{algorithmic}
\end{algorithm}

In the dual-level mixture comprehension (DLMC) stage, the MLLM backbone is fine-tuned under a next-token prediction objective to jointly reason about object-level and sensor-level tactile information, as detailed in Algorithm~\ref{alg:dlmc}. Two types of instructional prompts are designed to guide this process. For the object property description task, the model is instructed to describe the physical attributes of a contacted object—including roughness, hardness, and texture—based on tactile video tokens. A typical training sample follows the format ``\texttt{<T\_VID>} [tactile tokens] \texttt{</T\_VID>} Describe the physical properties of the contacted surface,'' with the expected completion such as ``The surface feels soft and slightly rough, with small bumpiness.'' For the sensor identification task, the model is trained to recognize the tactile sensor used to capture the data, thereby learning sensor-specific cues such as illumination color, marker layout, and gel elasticity. The corresponding prompt is structured as ``\texttt{<T\_VID>} [tactile tokens]\texttt{</T\_VID>} Identify which tactile sensor captured this video,'' with the expected answer like ``Captured by a GelSight Mini sensor.'' The overall comprehension loss combines these two objectives using $\mathcal{L}_{\text{DLMC}} = \mathcal{L}_{\text{prop}} + \lambda_{\text{sen}}\,\mathcal{L}_{\text{sen}},$where $\lambda_{\text{sen}}=0.1$ balances property and sensor supervision.

\begin{algorithm}[h]
\caption{Two-Stage Aligned Generation: Training}
\label{alg:two-stage-train}
\begin{algorithmic}[1]
\Require Pretrained touch encoder $\ETouch$, touch decoder $v_{\theta_2}(\cdot \mid t,F)$, multimodal encoder $\EMLLM$, DiT Projector $v_{\theta_1}(\cdot \mid t,F)$, sensor MLP adapter $\mathrm{MLP}_{\text{sen}}$; tactile queries $Q$; non-contact tactile image $V_i^{\text{sen}}$
\State \textbf{Stage I: Reconstruction}

\State $Z_i^{\text{sen}} \gets \ETouch(V_i^{\text{sen}})$ \Comment{encode non-contact state}

\State $Z_i^{\text{sen+obj}} \gets \ETouch(V_i^{\text{sen+obj}})$ \Comment{encode contact state}

\State \textbf{if} $\text{Bernoulli}(p_{\text{drop}})=1$ \textbf{then} $F_i^{\text{cond}} \leftarrow Z_i^{\text{sen}}$ 
\State \textbf{else} $F_i^{\text{cond}} \leftarrow Z_i^{\text{obj+sen}}$
\State Sample $t \sim \mathcal{U}(0,1)$, $z \sim \mathcal{N}(0,I)$; \quad $x_t \gets (1-t)z + tZ_i$; \quad $u_t^\ast \gets Z_i - z$
\State Minimize decoder flow loss:
\State \hspace{1.5em} $L_{\text{rec}} = \big\| v_{\theta_2}(x_t \mid t, F_i^{\text{cond}}) - u_t^\ast \big\|_2^2$
\State Update $\theta_2 \leftarrow \theta_2 - \eta \nabla_{\theta_2} L_{\text{rec}}$

\State \textbf{Stage II: Sensor-Aware Alignment}
\State $\hat{Q}_i \gets \EMLLM(T_i, Q)$ \Comment{textual/object-level features}
\State $S' \gets \mathrm{MLP}_{\text{sen}}(S)$ \Comment{project sensor tokens}
\State $F_i^{\text{obj+sen}} \gets [\hat{Q}_i; S']$
\State Sample $t \sim \mathcal{U}(0,1),\; z \sim \mathcal{N}(0,I)$
\State $x_t \gets (1-t)z + tZ_i$; \quad $u_t^{\ast} \gets Z_i - z$
\State Minimize $L_{\text{RF}} = \|v_{\theta_1}(x_t \mid t, F_i^{\text{obj+sen}}) - u_t^{\ast}\|_2^2$
\State Update $v_{\theta_1}$ by gradient descent on $L_{\text{RF}}$
\end{algorithmic}
\end{algorithm}

In parallel, the reconstruction stage jointly optimizes the touch encoder 
$\ETouch$ and decoder $\DTouch$ to reconstruct tactile images and videos from latent features, as described in Stage I of Algorithm~\ref{alg:two-stage-train}. This process enables the model to capture both object-level semantics and sensor-level configurations directly from tactile signals. For tactile image reconstruction, training runs for approximately 20 epochs with a batch size of 512 using the ZeRO-1 configuration. The initial learning rate is set to 1e-4, linearly warmed up for 5000 steps and decayed to 1e-5 by a cosine schedule. The optimization follows the standard flow matching loss formulation. For tactile video reconstruction, training is conducted with a batch size of 8, gradient accumulation set to 16, and gradient clipping enabled with a maximum norm of 1. This configuration employs the ZeRO-2 strategy for memory-efficient distributed optimization. The learning rate schedule and loss design are consistent with the image reconstruction setup, ensuring stable temporal consistency across consecutive frames. Together, these reconstruction settings allow UniTac to efficiently learn both static and dynamic tactile representations across multiple sensors while maintaining high-resolution fidelity and stable optimization behavior.

After these two stages converge, the NextDiT projector is trained for sensor-aware alignment, as shown in Stage II of Algorithm \ref{alg:two-stage-train}. The projector maps semantic embeddings from 
$\EMLLM$ into the tactile latent space produced by $\ETouch$. This stage runs for 100 epochs with a batch size of 512 and a learning rate of 1e-4, using the same warm-up strategy. The training objective employs the rectified flow matching loss.

All stages are trained under bf16 mixed precision. Experiments are conducted on a cluster equipped with 8 × NVIDIA A800 (80 GB) GPUs, managed by Accelerate for distributed optimization.

\subsection{Inference and Sampling}
\label{sec:inference}
At inference time, tactile generation proceeds through two sequential processes: the DiT Projector Sampling and the Sensor-Prior Sampling Strategy (SPSS), whose implementations are summarized in Algorithm~\ref{alg:projector-sample} and Algorithm~\ref{alg:decoder-spss}, respectively.

% =========================================
% Algorithm 2': DiT Projector — Normal Sampling (Inference, no SPSS)
% =========================================
\begin{algorithm}[h]
\caption{DiT Projector Sampling}
\label{alg:projector-sample}
\begin{algorithmic}[1]
\Require Trained DiT Projector $v_{\theta_1}$; multimodal encoder $\EMLLM$; sensor MLP $\mathrm{MLP}_{\text{sen}}$; text $T_i$; tactile queries $Q$; sensor tokens $S$; number of steps $N$
\State $\hat{Q}_i \gets \EMLLM(T_i, Q)$
\State $S' \gets \mathrm{MLP}_{\text{sen}}(S)$
\State $F_i^{\text{obj+sen}} \gets [\hat{Q}_i; S']$
\State Initialize $x^{(1)}_{1} \sim \mathcal{N}(0,I)$ \Comment{projector latent at $t{=}1$}
\For{$k = 1$ \textbf{to} $N$} \Comment{standard flow integration for $v_{\theta_1}$}
  \State $t \gets 1 - \frac{k-1}{N-1}$; \quad $\Delta \gets \frac{1}{N-1}$
  \State $\hat{v} \gets v_{\theta_1}(x^{(1)}_{t} \mid t, F_i^{\text{obj+sen}}))$
  \State $x^{(1)}_{t-\Delta} \gets \textsc{FlowStep}(x^{(1)}_{t}, \hat{v}, \Delta)$
\EndFor
\State $\tilde{Z}_i^{\text{obj+sen}} \gets x^{(1)}_{0}$ \Comment{projector’s aligned latent}
\State \Return $\tilde{Z}_i^{\text{obj+sen}}$
\end{algorithmic}
\end{algorithm}

% =========================================
% Algorithm 3: Sensor-Prior Sampling Strategy (Eq. 6–7)
% =========================================
\begin{algorithm}[h]
\caption{Touch Decoder with Sensor-Prior Sampling (SPSS)}
\label{alg:decoder-spss}
\begin{algorithmic}[1]
\Require Trained touch decoder $v_{\theta_2}$; touch encoder $\ETouch$; multimodal encoder $\EMLLM$; sensor MLP $\mathrm{MLP}_{\text{sen}}$; text $T_i$; non-contact tactile image $V_i^{\text{sen}}$; sensor tokens $S$; guidance scale $s$; number of steps $N$; projector’s aligned latent $\tilde{Z}_i^{\text{obj+sen}}$
\State $\hat{Q}_i \gets \EMLLM(T_i, Q)$; \quad $S' \gets \mathrm{MLP}_{\text{sen}}(S)$
\State $Z_i^{\text{sen}} \gets \ETouch(V_i^{\text{sen}})$ \Comment{encode non-contact state}

\State Initialize $x^{(2)}_{1} \sim \mathcal{N}(0, I)$ \Comment{decoder latent at $t{=}1$}
\For{$k = 1$ \textbf{to} $N$} \Comment{SPSS is applied inside $v_{\theta_2}$}
  \State $t \gets 1 - \frac{k-1}{N-1}$; \quad $\Delta \gets \frac{1}{N-1}$
  \State $v_{\text{sen}} \gets v_{\theta_2}(x^{(2)}_{t} \mid t, Z^{\text{sen}}_i)$
  \State $v_{\text{obj+sen}} \gets v_{\theta_2}(x^{(2)}_{t} \mid t, Z^{\text{obj+sen}}_i)$
  \State $\hat{v} \gets v_{\text{sen}} + s \cdot (v_{\text{obj+sen}} - v_{\text{sen}})$ \Comment{sensor-prior guidance (Eq.~6–7)}
  \State $x^{(2)}_{t-\Delta} \gets \textsc{FlowStep}(x^{(2)}_{t}, \hat{v}, \Delta)$
\EndFor
\State $\hat{Z}_i \gets x^{(2)}_{0}$; \quad $\hat{V}_i \gets v_{\theta_2}(\hat{Z}_i \mid t{=}0)$
\State \Return $\hat{V}_i$
\end{algorithmic}
\end{algorithm}

In the projector sampling stage, the trained NextDiT projector $v_{\theta_1}$ receives the text prompt $T_i$ and tactile queries $Q$, producing the aligned latent representation along the rectified-flow path. This step effectively transfers object-level semantics and sensor priors from the MLLM space into the tactile latent space defined by the encoder $\ETouch$. Subsequently, the sensor-prior sampling stage drives the touch decoder $v_{\theta_2}$ to synthesize realistic tactile signals that simulate the gradual transition from non-contact to contact. Unlike classifier-free guidance, SPSS introduces a sensor-conditioned branch $v_{\theta_2}(x_t \mid t, Z_i^{\text{sen}})$ representing the non-contact state, and a contact branch $v_{\theta_2}(x_t \mid t, Z_i^{\text{obj+sen}})$ reflecting object interaction. Their difference is scaled by a guidance factor $s = 1.5$ according to Eq.~(6)--(7), ensuring that the generated touch patterns remain physically consistent with real sensor behavior. Each generation trajectory uses 50 rectified-flow steps to progressively denoise the latent variable toward the final tactile frame or video.

\subsection{Real-World Deployment Experiment}
We validate UniTac in a real-world robotic deployment scenario to assess its performance in physical interaction tasks. The system is built upon a Tracer mobile base equipped with dual robotic arms and multi-sensor perception modules, including a tactile sensor Gelsight Mini mounted on the end-effector. All components are managed under the \texttt{cobot\_magic} ROS framework and run on an onboard industrial PC using ROS~Noetic. During deployment, the robot performs grasping on candidate materials, and streams tactile feedback to UniTac for real-time inference. In the fabric selection task, UniTac receives tactile sequences from two visually similar fabrics and, through its Property Comparison module, correctly infers that the smoother purple towel exhibits lower roughness. The robot then grasps this fabric as the preferred choice for baby skin contact, demonstrating UniTac’s capacity to integrate tactile understanding with physically grounded manipulation in real environments.

\section{More Experimental Results}
\label{sec:more_results}

\subsection{Does Tactile Feedback Improve Robotic Manipulation?}
\label{sec:appendix_robot_tactile_utility}

In this part, we evaluate whether tactile representations provide practical benefits for robotic manipulation. Rather than claiming that a vision-language-action (VLA) policy cannot solve contact-rich tasks, our goal is to examine whether tactile supervision improves contact-aware generalization and sample efficiency. We compare three policy variants: (1) \textbf{VLA} \cite{intelligence2025pi_}, which maps RGB observations, robot state, and language instructions to actions; (2) \textbf{VTLA-real}, which additionally uses real GelSight tactile observations during rollout; and (3) \textbf{VTLA-pred}, which does not use real-time tactile input at inference, but predicts a tactile representation from RGB inputs and conditions action prediction on this tactile latent. All settings are implemented based on $\pi_{0.5}$ \cite{intelligence2025pi_}.

\paragraph{Unseen-size cup deformation grasping.}
We first design a target deformation grasping task on paper cups. During training, the robot observes smaller cups with synchronized third-view RGB videos, first-view robot-camera videos, GelSight tactile images, and action trajectories. The task is to grasp each cup and compress it by a fixed target deformation. Specifically, the training cups correspond to smaller sizes, while evaluation is conducted on a larger unseen cup. The target deformation is set to $1$ cm, and a rollout is considered deformation-successful only when the final cup diameter lies within the target range.

\begin{figure}[t]
\centering
\includegraphics[width=0.95\linewidth]{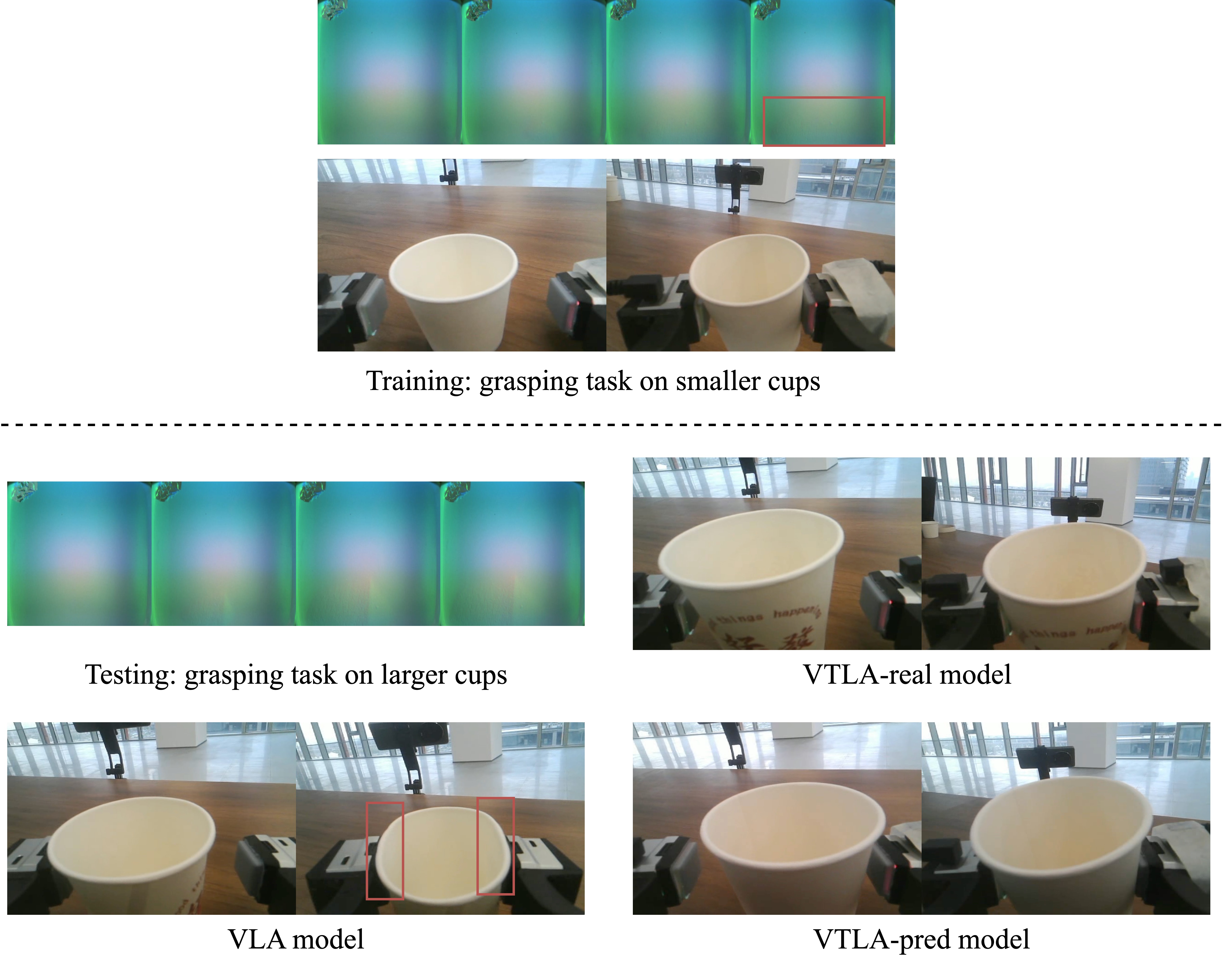}
\caption{Unseen-size target deformation grasping. The robot is trained on smaller cups with synchronized RGB and GelSight tactile observations, and tested on a larger unseen cup. RGB-only VLA can often lift the cup but result in unrealistic over-compression. VTLA with real tactile feedback achieves accurate deformation control, while VTLA with predicted tactile representation remains close to the real tactile condition.}
\label{fig:cup_deformation_grasp}
\end{figure}

This task separates general grasping from deformation-aware grasping. A policy may successfully lift the cup but still fail if it over-compresses or under-compresses the cup. As shown in Tab.~\ref{tab:cup_deformation_results}, the RGB-only VLA baseline achieves high lift success, indicating that visual observations are sufficient for localizing and grasping the cup in many cases. However, it fails to achieve the target deformation on the unseen larger cup. As shown in Fig.~\ref{fig:cup_deformation_grasp}, the VLA policy applies excessive gripping force on the larger cup, resulting in unrealistic over-compression and undesirable structural deformation. This suggests that the visual policy learns where to grasp but lacks direct access to the contact state needed for accurate force and deformation control. In contrast, VTLA-real achieves the best performance by directly observing GelSight deformation during contact. VTLA-pred is slightly worse than VTLA-real but remains substantially better than VLA, showing that predicted tactile representations preserve most of the contact-aware benefit even without real-time tactile input.

\begin{table}[t]
\centering
\caption{Unseen-size cup deformation grasping. The robot is trained on smaller cups and evaluated on a larger unseen cup. The task is not only to lift the cup but also to compress it to the target deformation.}
\label{tab:cup_deformation_results}
\setlength{\tabcolsep}{5mm}
\resizebox{0.9\linewidth}{!}{
\begin{tabular}{l|cccc}
\toprule
Method & Lift Success $\uparrow$ & Target Deform. Success $\uparrow$ & Overall Success $\uparrow$ \\
\midrule
VLA & 18/20 (90\%) & 0/20 (0\%) & 0/20 (0\%) \\
VTLA-real & 20/20 (100\%) & 20/20 (100\%) & 20/20 (100\%) \\
VTLA-pred & 19/20 (95\%) & 18/20 (90\%) & 18/20 (90\%) \\
\bottomrule
\end{tabular}}
\end{table}

\paragraph{Discussion.}
These two experiments demonstrate complementary roles of tactile representations in robotic manipulation. In the cup experiment, tactile observations provide a local deformation cue that is more directly related to contact force than global visual geometry, enabling better deformation control on unseen object sizes. In the towel experiment, tactile representations ground high-level language concepts such as softness and skin-friendliness while also revealing contact state, layer thickness, and slip. Overall, VLA primarily learns visual-action correlations, whereas VTLA learns whether the contact state itself is appropriate. This contact-aware representation enables more robust behavior under unseen object size, ambiguous visual appearance, and tactile-property-dependent instructions.

\subsection{Effect of $\lambda_{\text{sen}}$.}
We further analyze the influence of the weighting factor $\lambda_{\text{sen}}$ in the Dual-Level Mixture Comprehension loss, which balances the object-level property description and the sensor-level identification objectives.
As shown in Tab.~\ref{tab:lambda_ablation}, 
a small $\lambda_{\text{sen}}$ ($<0.05$) weakens the model’s sensitivity to sensor-specific cues, resulting in inconsistent tactile alignment across sensors. 
Conversely, a large value ($>0.5$) overemphasizes sensor discrimination, slightly degrading object-level property reasoning and reconstruction quality.
When $\lambda_{\text{sen}}=0.1$, UniTac achieves the best overall performance across PHYSICLEAR, SSIM, and PSNR, indicating an optimal trade-off between cross-sensor consistency and tactile understanding.

\begin{table}[h]
\centering
\caption{Ablation study on the weighting factor $\lambda_{\text{sen}}$.}
% \vspace{-0.3em}
\label{tab:lambda_ablation}
\setlength{\tabcolsep}{8mm}
\resizebox{0.75\linewidth}{!}{
\begin{tabular}{c|ccc}
\toprule
\textbf{$\lambda_{\text{sen}}$} & PHYSICLEAR & SSIM & PSNR \\
\midrule
0.01  & 58.27 & 0.818 & 19.42 \\
0.05 & 59.36 & 0.826 & 19.71 \\
0.1 & \textbf{60.61} & \textbf{0.836} & \textbf{19.93} \\
0.5  & 59.84 & 0.829 & 19.65 \\
1  & 58.12 & 0.817 & 19.28 \\
\bottomrule
\end{tabular}
}
\vspace{-0.5em}
\end{table}

\subsection{Touch Encoder Choice}

We further investigate which image representation is more suitable for the touch encoder $\ETouch$: a low-level pixel representation extracted from a VAE-based encoder that preserves detailed spatial textures, or a high-level semantic representation extracted from a CLIP-style vision encoder that captures abstract visual cues. Both variants are trained following the same procedure described in Sec.~\ref{sec:training}, using identical decoder and projector settings.

\begin{table}[h]
\centering
\caption{Comparison between low-level pixel features and high-level semantic features as tactile image representations. 
All results are measured under identical training and inference settings.}
% \vspace{-0.3em}
\label{tab:rep_ablation}
\setlength{\tabcolsep}{5mm}
\resizebox{1\linewidth}{!}{
\begin{tabular}{lcccc}
\toprule
Representation Type & Encoder Source & PHYSICLEAR & SSIM & PSNR \\
\midrule
Low-Level Pixel & VAE-based & 60.55 & 0.836 & 19.95 \\
High-Level Semantic & CLIP-based & 60.61 & 0.836 & 19.93 \\
\bottomrule
\end{tabular}
}
% \vspace{-0.5em}
\end{table}

Tab.~\ref{tab:rep_ablation} indicates that both representations achieve comparable tactile reconstruction quality, with only marginal differences across PHYSICLEAR, SSIM, and PSNR metrics. However, the VAE-based pixel encoder produces dense features and incurs a substantially higher inference latency. In contrast, the CLIP-based semantic encoder yields compact low-Dimension embeddings that are both computationally efficient and semantically aligned with language supervision, which benefits cross-modal reasoning in the Dual-Level Mixture Comprehension stage. Therefore, we adopt the semantic representation as the default encoder input, striking an effective balance between fidelity, efficiency, and multimodal alignment.

\subsection{Number of Queries for \texorpdfstring{$\EMLLM$}{EMLLM}}
We ablate the number of textual–tactile queries used by the MLLM. All settings follow the same training schedule as described in Sec.~\ref{sec:training} and use the same inference pipeline as in Sec.~\ref{sec:inference}. As shown in Tab.~\ref{tab:query_ablation}, increasing the number of queries from 16 to 64 consistently improves the generation metrics, whereas further scaling to 128 or 256 yields only marginal gains (\(\leq\!0.03\) PSNR) while introducing additional computational overhead. Considering the trade-off between accuracy and efficiency, we adopt 64 queries as the default setting.

\begin{table}[t]
\centering
\caption{Ablation on the number of queries fed into the MLLM.}
% \vspace{-0.35em}
\label{tab:query_ablation}
\setlength{\tabcolsep}{15mm}
\resizebox{0.8\linewidth}{!}{
\begin{tabular}{c|cccc}
\toprule
\#Queries & SSIM & PSNR \\
\midrule
16 & 0.820 & 19.08 \\
32 & 0.824 & 19.76 \\
64 & \textbf{0.836} & \textbf{19.93} \\
128 & 0.836 & 19.95 \\
256 & 0.836 & 19.96 \\
\bottomrule
\end{tabular}
}
% \vspace{-0.6em}
\end{table}

\subsection{Effect of Sensor-prior Sampling}
In Tab.~\ref{tab:ablation}, the variant ``w/o sensor-prior sampling'' degenerates to vanilla classifier-free guidance (CFG), where the unconditional branch is used for guidance. To disentangle the effect of the guidance scale from the proposed sensor-prior sampling strategy, we further conduct a guidance-scale ablation in Tab.~\ref{tab:spss_scale}. Vanilla CFG achieves its best performance at $s=1.5$, with 0.817 SSIM and 19.49 PSNR. Under the same guidance scale, SPSS improves the results to 0.836 SSIM and 19.93 PSNR. These results indicate that the improvement of SPSS is not merely brought by guidance scaling, but mainly comes from replacing the unconditional branch with the sensor-conditioned prior, which provides more informative tactile guidance during generation.

\begin{table}[H]
\centering
\caption{Ablation study on the guidance scale $s$ of SPSS. The matched comparison with vanilla CFG is conducted at $s=1.5$, where CFG achieves its best performance.}
\label{tab:spss_scale}
\setlength{\tabcolsep}{5mm}
\resizebox{1\linewidth}{!}{
\begin{tabular}{c|c|cccccc}
\toprule
Metric
& CFG, $s=1.5$
& SPSS, $s=1.0$
& $s=1.5$
& $s=2.0$
& $s=3.0$
& $s=5.0$
& $s=7.5$ \\
\midrule
SSIM
& 0.817 & 0.804 & \textbf{0.836} & 0.830 & 0.822 & 0.806 & 0.791 \\
PSNR
& 19.49 & 19.51 & \textbf{19.93} & 19.82 & 19.46 & 19.07 & 18.54 \\
\bottomrule
\end{tabular}}
\end{table}

\subsection{Additional Cross-Sensor Grasp Classification Results}

To further examine whether the cross-sensor generalization gain is specific to the Digit$\rightarrow$GelSight setting, we conduct additional grasp classification experiments on two transfer directions: GelSight$\rightarrow$Duragel and Duragel$\rightarrow$Digit. As shown in Tab.~\ref{tab:supp_cross_sensor_grasp}, the source-only classifier performs well on the source sensor but suffers a substantial accuracy drop on the target sensor, indicating a clear sensor-domain shift. By augmenting the source data with UniTac-generated target samples, the target-domain accuracy is greatly improved while the source-domain accuracy is largely preserved. The resulting performance approaches the real-target upper bound, further demonstrating the effectiveness of UniTac for cross-sensor tactile data generation and downstream transfer.

\begin{table}[h]
\centering
\caption{Additional cross-sensor grasp classification accuracy with generated target data. For each transfer direction, we report the classification accuracy on both the source and target sensors.}
\label{tab:supp_cross_sensor_grasp}
\setlength{\tabcolsep}{5mm}
\resizebox{0.8\linewidth}{!}{
\begin{tabular}{l|cc|cc}
\toprule
\multirow{2}{*}{Training Data}
& \multicolumn{2}{c|}{GelSight$\rightarrow$Duragel}
& \multicolumn{2}{c}{Duragel$\rightarrow$Digit} \\
\cmidrule(lr){2-3} \cmidrule(lr){4-5}
& GelSight & Duragel
& Duragel & Digit \\
\midrule
Source only
& 98.84 & 53.11
& 97.89 & 52.46 \\
Source + UniTac target
& 96.12 & 94.48
& 96.02 & 95.75 \\
Source + Real-target
& 97.35 & 96.20
& 96.73 & 97.42 \\
\bottomrule
\end{tabular}}
\end{table}

\section{Additional Qualitative Results}
\label{sec:qualitative}

\subsection{Reconstruction Results}
\begin{figure}[t]
  \centering
   \includegraphics[width=1.0\linewidth]{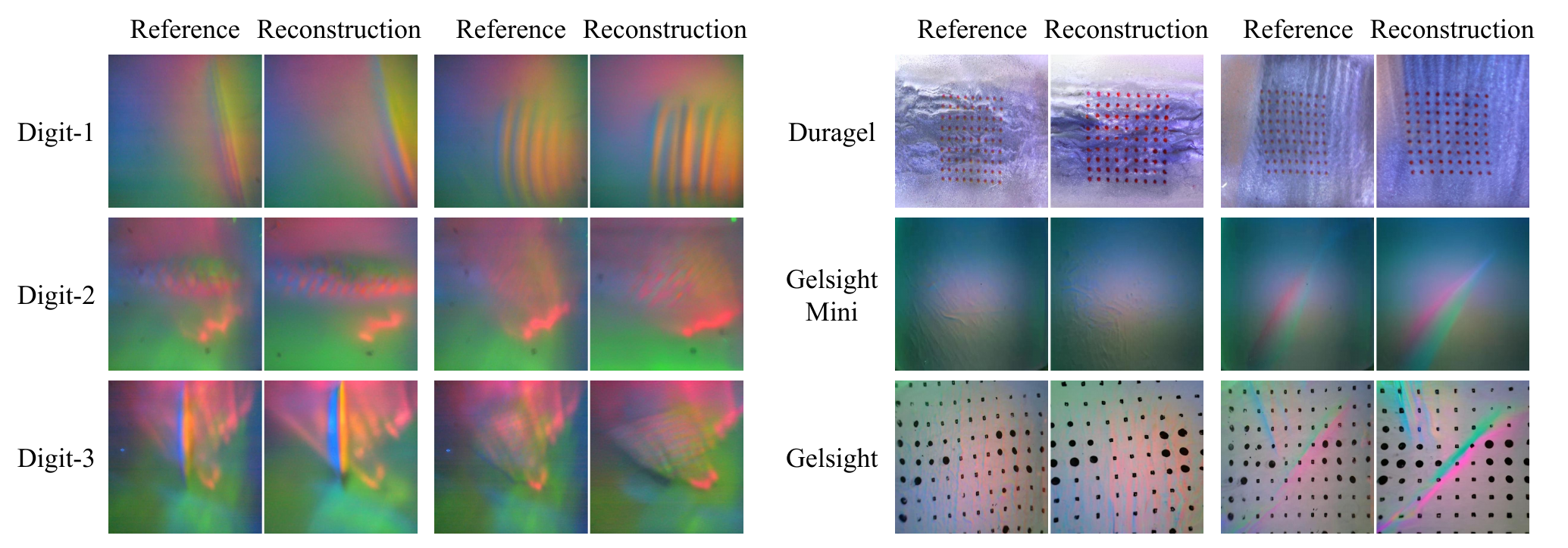}
    % \vspace{-2em} 
   \caption{Reconstruction results after training the touch decoder reconstruction task.}
   \label{fig:reconstruction}
    % \vspace{-3mm} 
\end{figure}

Fig.~\ref{fig:reconstruction} illustrates the reconstruction quality of UniTac across several tactile sensors, including Digit, Duragel, GelSight Mini and GelSight. For the Digit sensors, the reconstructed frames remain highly faithful to the reference images, capturing both the characteristic color fringes and the fine-grained deformation patterns. GelSight and GelSight Mini also exhibit strong consistency between reference and reconstruction, with clear preservation of grid alignment, illumination gradients and surface-normal cues.

In contrast, the Duragel reconstructions appear less stable. This is consistent with the nature of the original Duragel data, which contains fluctuations during collection and often presents weaker or noisier optical responses. Despite this challenge, the decoder still recovers the main structural patterns, but the results show more variation compared with other sensors.

Overall, the reconstruction stage successfully learns a reliable and sensor-aware latent representation. The model performs consistently across most tactile modalities, and the reduced performance on Duragel reflects the inherent instability of the raw data rather than limitations of the training framework.

\subsection{Additional Generation and Understanding Results.}
\begin{figure}[t]
  \centering
   \includegraphics[width=1.0\linewidth]{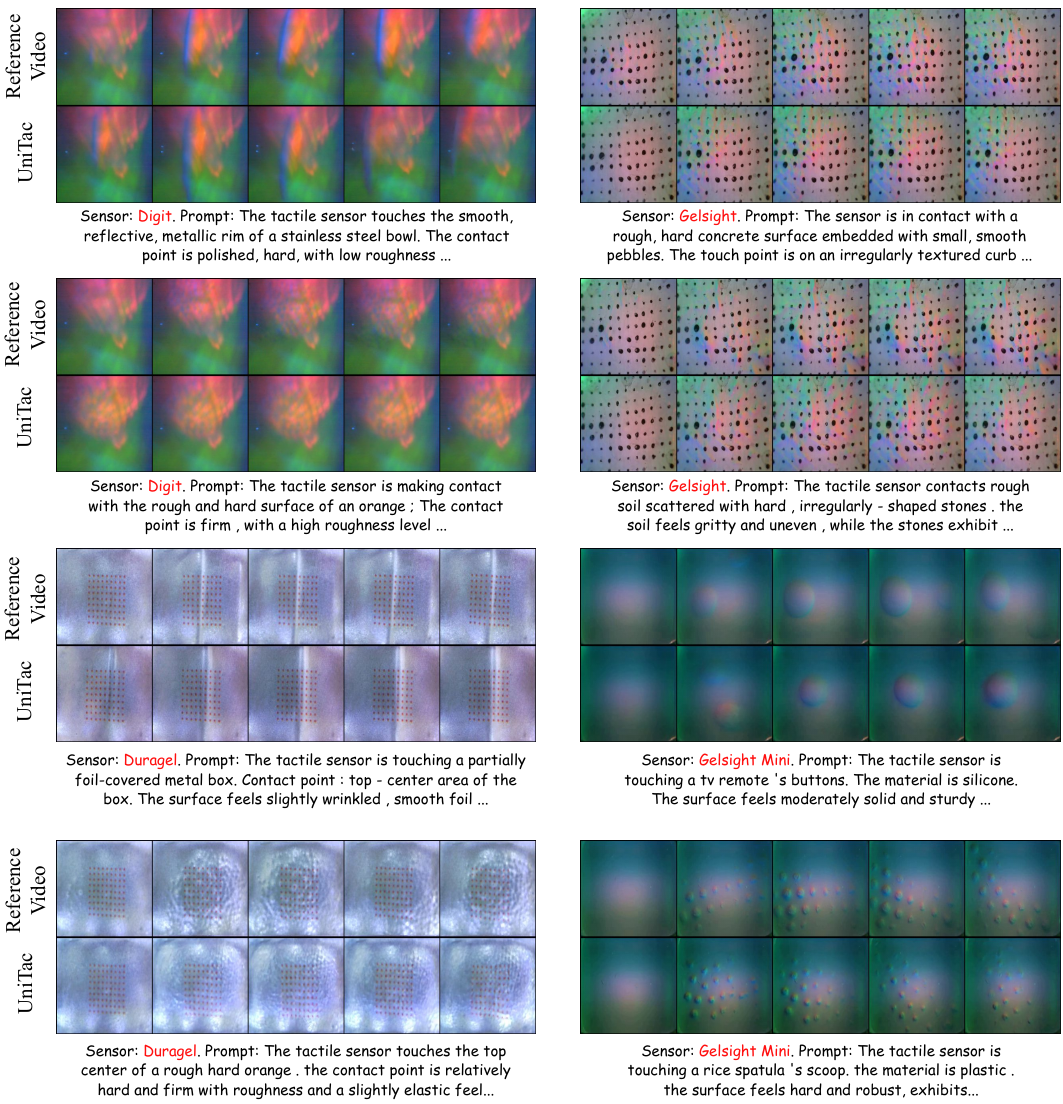}
    \vspace{-2em} 
   \caption{More results on generating tactile videos using various tactile sensors.}
   \label{fig:touch_video_generation_res_appendix}
    % \vspace{-3mm} 
\end{figure}

Fig.~\ref{fig:touch_video_generation_res_appendix} presents additional tactile video generation results across Digit, Duragel, GelSight, and GelSight Mini sensors. UniTac produces coherent temporal dynamics and preserves sensor-specific appearance, such as the characteristic color fringes of Digit and the structured grid patterns of GelSight. The generated contact sequences remain consistent with the textual prompts, demonstrating that the learned cross-sensor latent space effectively supports conditional tactile synthesis.

\begin{figure}[t]
  \centering
   \includegraphics[width=1.0\linewidth]{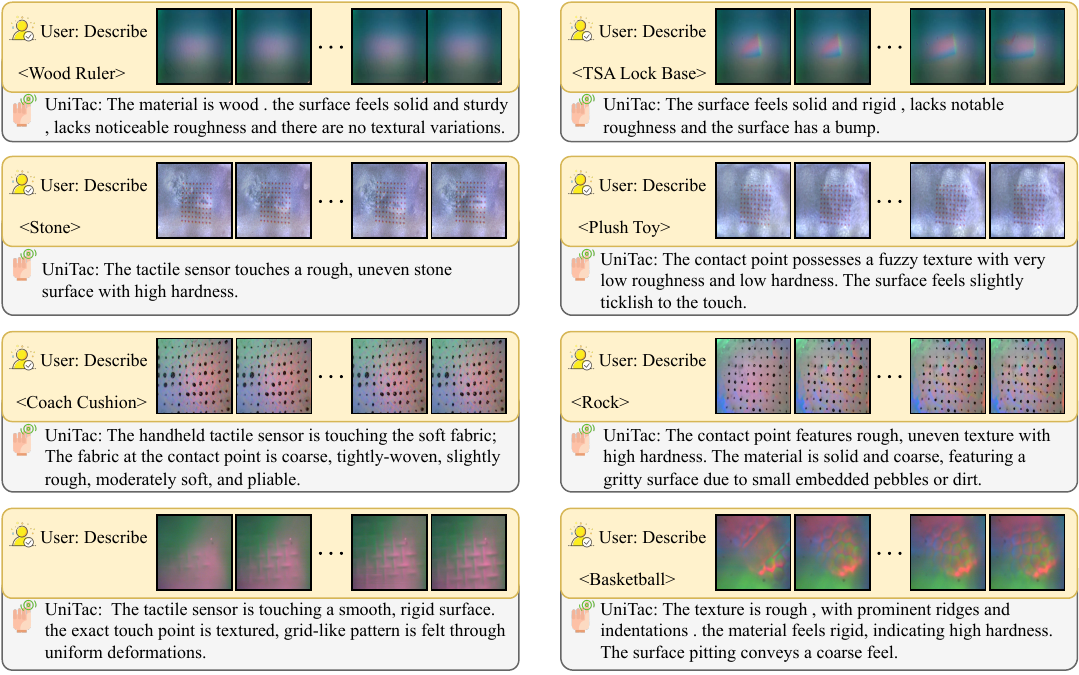}
    \vspace{-2em} 
   \caption{More results on understanding tactile videos using various tactile sensors.}
   \label{fig:touch_video_understanding_res_appendix}
    % \vspace{-3mm} 
\end{figure}

Fig.~\ref{fig:touch_video_understanding_res_appendix} provides further qualitative examples for tactile understanding. Given tactile video clips from different materials, UniTac generates concise and semantically aligned descriptions that reflect surface hardness, roughness, texture patterns, and deformation behavior. These examples confirm that UniTac generalizes well across various material categories and maintains stable understanding performance, complementing the quantitative evaluations reported in the main paper.

\section{Preliminaries of Rectified Flow Matching}
\label{sec:preliminaries}

Rectified flow models provide an alternative generative formulation that avoids the stochastic noise injection used in diffusion probabilistic models. Instead of constructing a Markov chain that gradually perturbs a data sample with Gaussian noise, rectified flow defines a deterministic transport path between a simple prior distribution and the target data distribution and learns a velocity field that moves samples along this path. This results in a simplified training objective and an efficient deterministic sampling procedure.

Given a data point $\mathbf{x}_{0} \sim q_{\text{data}}(\mathbf{x})$ and a noise sample $\mathbf{z} \sim \mathcal{N}(0,I)$, rectified flow introduces a linear interpolation connecting these two endpoints:
\begin{equation}
    \mathbf{x}_{t} = (1 - t)\,\mathbf{z} + t\,\mathbf{x}_{0}, \qquad t \in [0,1].
\end{equation}
The instantaneous velocity along this path is given by
\begin{equation}
    \mathbf{u}_{t}^{\ast} = \frac{\mathrm{d}\mathbf{x}_{t}}{\mathrm{d}t} = \mathbf{x}_{0} - \mathbf{z}.
\end{equation}
A neural network $\mathbf{u}_{\theta}(\mathbf{x}_{t}, t)$ is then trained to approximate this ground-truth velocity by minimizing the flow matching loss:
\begin{align}
\mathcal{L}_{\text{FM}}
    &= \mathbb{E}_{t,\,\mathbf{x}_{0},\,\mathbf{z}}
    \left[\,\left\| \mathbf{u}_{\theta}(\mathbf{x}_{t}, t) - (\mathbf{x}_{0} - \mathbf{z}) \right\|_{2}^{2} \right].
\end{align}
Unlike diffusion models, rectified flow does not require noise schedules or variance-preserving constraints, and the linear path structure makes the training objective straightforward and stable.

At inference time, sample generation proceeds by solving a first-order ordinary differential equation defined by the learned velocity field:
\begin{equation}
    \frac{\mathrm{d}\mathbf{x}_{t}}{\mathrm{d}t} = \mathbf{u}_{\theta}(\mathbf{x}_{t}, t),
\end{equation}
initialized from a Gaussian latent $\mathbf{x}_{1} \sim \mathcal{N}(0,I)$.  
A discrete solver with $N$ integration steps applies updates of the form
\begin{equation}
    \mathbf{x}_{t_{k+1}} = \mathbf{x}_{t_{k}} - \Delta t \cdot \mathbf{u}_{\theta}(\mathbf{x}_{t_{k}}, t_{k}),
\end{equation}
where $t_{k} = 1 - \frac{k-1}{N-1}$ and $\Delta t = \frac{1}{N-1}$.  
This deterministic integration transports the sample smoothly toward the target distribution.

To enable conditional generation conditioned on auxiliary inputs $\mathbf{c}$ such as textual prompts or sensor-related features, classifier-free guidance is integrated into the rectified flow formulation. Let $\mathbf{u}_{\theta}(\mathbf{x}_{t}, t, \mathbf{c})$ denote the conditional velocity and $\mathbf{u}_{\theta}(\mathbf{x}_{t}, t, \emptyset)$ denote the unconditional velocity. The guided velocity is computed as
\begin{equation}
\tilde{\mathbf{u}}_{\theta}(\mathbf{x}_{t}, t, \mathbf{c})
    = s \cdot \mathbf{u}_{\theta}(\mathbf{x}_{t}, t, \mathbf{c})
      + (1 - s) \cdot \mathbf{u}_{\theta}(\mathbf{x}_{t}, t, \emptyset),
\end{equation}
where $s$ is the guidance scale. This approach enhances the influence of the conditional signal without requiring an explicit classifier and is directly compatible with the ODE-based sampling process.

Through this combination of a simple training objective, deterministic sampling dynamics, and classifier-free conditioning, rectified flow provides an efficient and flexible framework for generative modeling. In our work, this formulation plays a key role in both the sensor-aware projector and the tactile decoder, enabling stable multimodal alignment and physically coherent tactile synthesis.

\end{document}